\documentclass[twoside]{article}

\usepackage[preprint]{aistats2027}
\usepackage{hyperref}
\usepackage[numbers]{natbib}

\usepackage{amssymb}            
\usepackage{mathtools}          
\usepackage{mathrsfs}           
\usepackage{graphicx}           
\usepackage{subcaption}         
\usepackage[space]{grffile}     
\usepackage{url}                
\usepackage{lipsum}             
\usepackage{amsmath}
\usepackage{amssymb}
\usepackage{mathtools}
\usepackage{amsthm}

\usepackage{xfrac}
\usepackage{subcaption}
\usepackage{graphicx}
\usepackage{pgfplots}

\usepackage{algorithm}
\usepackage[noend]{algorithmic}
\renewcommand\algorithmiccomment[1]{%
  \hfill {(#1)}%
}

\usepackage{wrapfig}
\usepackage{filecontents} 
\usepackage{enumitem} 
\usepackage{adjustbox}
\usepackage{arydshln} 

\usepackage{tikz}
\usetikzlibrary{automata, positioning, arrows.meta, calc, fit, backgrounds}
\usepackage{thm-restate}
\usepackage{dsfont}
\usepackage{bbm}
\usepackage{enumitem} 
\usepackage{mathtools}
\usepackage[textsize=tiny]{todonotes}
\usepackage{booktabs}       
\usepackage{xcolor}

\theoremstyle{plain}
\newtheorem{theorem}{Theorem}[section]
\newtheorem{proposition}[theorem]{Proposition}
\newtheorem{lemma}[theorem]{Lemma}
\newtheorem{corollary}[theorem]{Corollary}
\theoremstyle{definition}
\newtheorem{definition}[theorem]{Definition}
\newtheorem{assumption}[theorem]{Assumption}
\theoremstyle{remark}
\newtheorem{remark}[theorem]{Remark}

\newcommand{\vic}[1]{\todo[color=blue!10, inline]{\small VV: #1}}
\newcommand{\tf}[1]{\todo[color=red!10, inline]{\small TF: #1}}
\newcommand{\andreas}[1]{\todo[color=green!10, inline]{\small AS: #1}}
\definecolor{darkgreen}{RGB}{0,110,50}

\DeclareMathOperator{\EX}{\mathbb{E}}
\newcommand{\R}{\mathbb{R}}
\DeclareMathOperator*{\argmax}{arg\,max}
\DeclareMathOperator*{\argmin}{arg\,min}
\newcommand{\floor}[1]{\left \lfloor #1 \right \rfloor }
\newcommand{\roof}[1]{\left \lceil #1 \right \rceil }
\newcommand*{\defeq}{\stackrel{\Delta}{=}}
\newcommand{\bc}[1]{\left\{{#1}\right\}}
\newcommand{\br}[1]{\left({#1}\right)}
\newcommand{\bs}[1]{\left[{#1}\right]}
\newcommand{\abs}[1]{\left| {#1} \right|}
\newcommand{\norm}[1]{\left\lVert#1\right\rVert}
\newcommand\ip[2]{\langle #1, #2 \rangle}
\newcommand{\ones}{\bm 1}
\newcommand{\ind}{\mathbbm{1}}
\newcommand{\diff}{\mathop{}\!\mathrm{d}}
\newcommand{\KL}[2]{\operatorname{KL}\!\left(#1 \,\|\, #2\right)}
\newcommand{\im}{\operatorname{im}}
\newcommand{\var}{\operatorname{Var}}
\newcommand{\cov}{\operatorname{Cov}}
\newcommand{\tr}{\operatorname{tr}}
\newcommand{\proj}{\operatorname{proj}}
\newcommand{\rest}[2]{\left. #1 \right|_{#2}}

\newcommand{\cA}{\mathcal{A}}
\newcommand{\cB}{\mathcal{B}}
\newcommand{\cC}{\mathcal{C}}
\newcommand{\cD}{\mathcal{D}}
\newcommand{\cE}{\mathcal{E}}
\newcommand{\cF}{\mathcal{F}}
\newcommand{\cG}{\mathcal{G}}
\newcommand{\cH}{\mathcal{H}}
\newcommand{\cI}{\mathcal{I}}
\newcommand{\cJ}{\mathcal{J}}
\newcommand{\cK}{\mathcal{K}}
\newcommand{\cL}{\mathcal{L}}
\newcommand{\cM}{\mathcal{M}}
\newcommand{\cN}{\mathcal{N}}
\newcommand{\cO}{\mathcal{O}}
\newcommand{\cP}{\mathcal{P}}
\newcommand{\cQ}{\mathcal{Q}}
\newcommand{\cR}{\mathcal{R}}
\newcommand{\cS}{\mathcal{S}}
\newcommand{\cT}{\mathcal{T}}
\newcommand{\cU}{\mathcal{U}}
\newcommand{\cV}{\mathcal{V}}
\newcommand{\cW}{\mathcal{W}}
\newcommand{\cX}{\mathcal{X}}
\newcommand{\cY}{\mathcal{Y}}
\newcommand{\cZ}{\mathcal{Z}}

\newcommand{\states}{\mathcal{S}}
\newcommand{\actions}{\mathcal{A}}
\newcommand{\dyns}{\mathcal{P}}
\newcommand{\dyn}{P}
\newcommand{\bdyn}{\mathbf{P}}

\newcommand{\pr}{\mathbb{P}}
\newcommand{\mdps}{\mathcal{M}}
\newcommand{\mdp}{\mu}
\newcommand{\ent}{\mathcal{H}}

\newcommand{\rE}{{r^\text{E}}}
\newcommand{\piE}{\pi^\text{E}}
\newcommand{\muE}{\mu^\text{E}}

\newcommand{\bpi}{\boldsymbol{\pi}}

\newcommand{\tE}{{{\theta}^\text{E}}}
\renewcommand{\t}{{{\theta}}}
\newcommand{\T}{{{\Theta}}}

\newcommand{\bphi}{{\boldsymbol{\phi}}}
\newcommand{\nuE}{\nu^\text{E}}

\newcommand{\tauE}{{\tau^\text{E}}}
\newcommand{\qE}{q^\text{E}}
\newcommand{\rmax}{r_\text{max}}
\newcommand{\rmin}{r_\text{min}}
\newcommand{\HE}{H^\text{E}}
\newcommand{\kE}{\kappa^\text{E}}
\newcommand{\pimin}{\pi_{\text{min}}}

\newcommand{\subopt}{\operatorname{SubOpt}}

\begin{document}

%

%
\runningauthor{Villin, Freihaut, Schlaginhaufen, Dimitrakakis, Kamgarpour, Ramponi}

\twocolumn[
\aistatstitle{Inferring Transferable Rewards via Active Inverse Reinforcement Learning}

\aistatsauthor{ Victor Villin \And Till Freihaut$^\dagger$ \And Andreas Schlaginhaufen$^\dagger$ }
\aistatsaddress{ University of Neuchâtel \And University of Zurich \And EPFL }

\aistatsauthor{ Christos Dimitrakakis$^*$ \And Maryam Kamgarpour$^*$ \And Giorgia Ramponi$^*$ }
\aistatsaddress{ University of Neuchâtel \And EPFL \And University of Zurich }
]

{\renewcommand{\thefootnote}{\fnsymbol{footnote}}
 \footnotetext[2]{Equal contribution.}
 \footnotetext[1]{Authors ordered alphabetically.}}

\begin{abstract}
Inverse Reinforcement Learning (IRL) aims to recover expert rewards that transfer across environments with differing dynamics. Prior work has shown that such transfer is possible when gathering expert demonstrations across sufficiently different dynamics. In practice, however, expert demonstrations are costly, making it critical to select the right dynamics to reduce the quantity of expert data required for learning transferable rewards.
We propose a two-phase approach: we first cycle through the available dynamics to obtain a reliable reward estimate, then actively select dynamics to reduce our uncertainty about the expert reward. Crucially, this yields novel sample complexity guarantees on transfer that hold for any dynamics, with a rate independent of the number of available dynamics. Empirically, our method significantly reduces the number of expert demonstrations needed to learn a transferable reward.


\end{abstract}

\section{Introduction}\label{sec:intro}

Learning a \emph{transferable} representation of expert behavior makes it possible to predict how the expert would act in previously unseen circumstances. Inverse Reinforcement Learning (IRL) \citep{ng2000algorithms} formalizes this by inferring a reward function from expert demonstrations. Unlike imitation learning methods such as Behavior Cloning (BC) \citep{NIPS1988_812b4ba2}, which directly replicate observed behavior, IRL aims to recover the underlying objective the expert is optimizing. In contrast to a cloned policy tied to its training dynamics, a reward can, in principle, transfer to unseen dynamics.

Despite its appeal, IRL is fundamentally ill-posed~\citep{ng2000algorithms}. 
Under entropy regularization~\citep{geist2019theory}, the standard setting in identifiability works~\citep{cao2021identifiability,rolland2022identifiability}, demonstrations from a fixed Markov decision process (MDP) identify the reward only up to subspace specific to that MDP.
 Consequently, such rewards transfer only to dynamics whose identifiable subspace sufficiently overlaps with that of the original MDP~\citep{schlaginhaufen2024towards}. Recent works show that observing expert demonstrations in as few as two sufficiently different dynamics resolves this ambiguity~\citep{ cao2021identifiability, rolland2022identifiability, schlaginhaufen2024towards}. However, these works leave open how to actively select such dynamics to reduce the ambiguity with as few expert demonstrations as possible.

In practice, expert demonstrations are costly, and only a fixed set of dynamics is available. We must therefore choose carefully which dynamics to query the expert in. This raises the following central question:
\begin{center}
\emph{How should we select dynamics to learn a transferable reward from a minimal number of expert demonstrations?}
\end{center}
To address this question, we propose an \emph{active} IRL setting, in which the learner chooses the dynamics under which the expert demonstrates the task. This setting is natural whenever reconfiguring the environment is cheap compared to collecting expert demonstrations. For example, a robotics lab can cheaply vary the initial placement, mass, or friction of manipulated objects~\citep{peng2018sim}, or a driving simulator can vary road conditions~\citep{dosovitskiy2017carla}; collecting demonstrations, by contrast, requires a skilled human operator for each one~\citep{mandlekar2018roboturk}.

Closest to our work, \citet{buening2024environment} adaptively design dynamics to identify the expert's reward. They empirically show that, compared to randomly chosen environments, this yields more transferable rewards from fewer demonstrations, but provide no theoretical guarantees. To the best of our knowledge, our work provides the first active IRL algorithms with sample complexity guarantees for learning transferable rewards through dynamics selection.

\textbf{Contributions.}
Our main contributions are:
\begin{enumerate}[itemsep=-2pt, topsep=-2pt, leftmargin=13pt]
    \item We formalize active IRL as a sequential problem: over $n$ rounds, the learner chooses dynamics from a finite set on which to request an expert demonstration, aiming to recover a reward that transfers as much as possible across the set (Equation~\eqref{eq:objective}).
    
    \item We derive novel confidence sets for the expert reward (Lemma~\ref{thm:confidencenew}) and show that transfer to each dynamics is controlled by how informative they are relative to the information already accumulated (Lemma~\ref{lemma:transfer}).

    \item This yields a two-phase algorithm: cycle through the set of dynamics until the reward estimate is reliable (Lemma~\ref{lemma:passive_learning}), then query the expert where information gain is greatest. After $n$ rounds of dynamics selection and expert demonstrations, this results in a transfer error of $\tilde\cO(1/n)$ across the set of available dynamics, independent of its cardinality (Theorem~\ref{theo:active_learning}). The same rate extends to \emph{any} dynamics, up to an irreducible bias 
    (Theorem~\ref{theorem:transfer}).

    \item We validate our approach empirically against imitation learning and IRL baselines, as well as dynamics design for IRL~\citep{buening2024environment}. Our active approach learns an equally transferable reward using up to five times fewer demonstrations than the baselines, and its learned reward generalizes best to unseen dynamics (Section~\ref{sec:experiments}).
\end{enumerate}

\section{Problem formulation}
\label{sec:formulation}

\textbf{Notations.}
Let $\norm{\cdot}_2$ denote the Euclidean norm. For a symmetric positive semidefinite matrix $X\in\R^{d\times d}$, $\|X\|_2$ denotes its spectral norm, 
$X^\dagger$ its Moore-Penrose inverse, and $\|v\|_X \coloneq \sqrt{\langle v, Xv \rangle}$ the associated Mahalanobis norm of a vector $v\in\R^d$. For a finite set $\cZ$, $\Delta^{\cZ}$ denotes the probability simplex over $\cZ$.

\textbf{Regularized MDPs.} A regularized MDP is a tuple $\cM =(\cS, \cA, T, \nu_1, r, H, \eta)$. Here, $\cS$ and $\cA$ are finite state and action spaces, $T: \cS \times \cA \to \Delta^\cS$ is the transition function, $\nu_1 \in \Delta^\cS$ denotes the initial state distribution, $r: \cS\times \cA \to \R$ denotes the reward, and $H \in \mathbb{N}^+$ is the horizon. We consider entropy-regularized MDPs, with $\eta>0$ denoting the inverse temperature that controls the strength of the regularization. Starting from state $s_1 \sim \nu_1$, an agent interacting with the MDP takes at each timestep $h$ an action $a_h \in\cA$, receives the reward $r(s_h, a_h)$ and the environment transitions to a new state $s_{h+1} \sim T(\cdot \mid s_h, a_h)$. Denoting the dynamics by $P = (\nu_1, T)$, the optimal solution of $\cM$ is a policy $\pi = (\pi_h)_{h=1}^H$, with $\pi_h:\mathcal{S}\to\Delta^{\mathcal{A}}$, maximizing the regularized value $V^\pi_{P}(r) \coloneq \mathbb{E}_{P}^\pi\bs{ \sum_{h=1}^{H} r(s_h,a_h) - \eta^{-1} \log\br{\pi_h(a_h \mid s_h)}}$. A state $s$ is unreachable at timestep $h$ if no policy visits it with positive probability at $h$. As a convention, all policies follow an arbitrarily fixed action distribution (e.g., uniform) on unreachable states, which leaves trajectory laws and values unchanged. The MDP $\cM$ then admits a unique optimal policy~\citep{geist2019theory}. For a reward $r$ and dynamics $P$, we write $V^\star_{P}(r)$ and $\pi_P(r)$ for the corresponding optimal value and policy.

\textbf{Setting.}
Given an unknown reward $\rE$, an expert follows the optimal policy $\pi^\text{E}_P \coloneq \pi_P(\rE)$ under any chosen dynamics $P$. We consider a fixed, finite family of MDPs $\{\cM_i\}_{i}$ that share $\rE$ but differ in their dynamics $\cP := \{P_i\}_{i}$.  A demonstration in $\cM_i$ is a trajectory $\tau = \br{(s_h, a_h)}_{h=1}^H$ generated by $\pi^\text{E}_{P_i}$.

We consider an \emph{active} setting in which a learner selects dynamics sequentially, and an expert is queried after each selection. Over a budget of $n\in\mathbb{N}^+$ rounds, the learner selects at each round $k$ dynamics $P_k \in \cP$, and observes a batch of $m$ expert trajectories generated under $P_k$, for a total of $nm$ trajectories, gathering them in a dataset
$
    \cD^\text{E} \coloneq \bigcup_{k=1}^n \bc{\tau_{k,j}}_{j=1}^m.
$

\textbf{Objective.}
Given this budget, our goal is to select dynamics $\br{P_k}_{k=1}^n$ from $\cP$ and estimate a reward $r$ from $\cD^\text{E}$ that is \emph{transferable} across $\cP$. We quantify transferability by the worst-case suboptimality,
\begin{equation}
    \label{eq:objective}
    \max_{P\in\cP}\; \subopt(r, P) \;=\; \max_{P\in\cP} V^\star_{P}(\rE) - V^{\pi_P(r)}_{P}(\rE),
\end{equation}
and we seek a reward $r$ minimizing~\eqref{eq:objective}. This objective captures the \emph{ground-truth} performance of the recovered optimal policy $\pi_P(r)$ across $\cP$ and coincides with the notion of transferability studied by~\citet{schlaginhaufen2024towards} for fixed dynamics. Furthermore, it can be viewed as a regularized analogue of the imitation gap studied in imitation learning~\citep{swamy2021moments}.

\textbf{Structural assumption.}
Before presenting the algorithm for selecting the dynamics $P_k$ and estimating the reward, we introduce our main structural assumption, which enables efficient estimation of the expert reward $\rE$ from expert demonstrations $\cD^\text{E}$.
\begin{assumption}[Expert bounded linear rewards]
    \label{as:linear_rewards}
    Let $\phi : \cS \times \cA \to \R^{d}$ be a known feature map with $\|\phi(s,a)\|_2 \le B_\phi$ for all $(s,a) \in \cS \times \cA$, and let $\T \coloneq \{\t \in \R^d : \|\t\|_2 \le B_\t\}$, where $B_\phi, B_\t > 0$ are known constants. We assume the expert reward is linear in \emph{unknown} parameters $\tE \in \T$, i.e.\ $\rE(s,a) = \ip{\tE}{\phi(s,a)}$.
\end{assumption}
Linear rewards are standard in RL~\citep{pmlr-v125-jin20a} and inverse RL~\citep{ng2000algorithms}. We index rewards $r_\t = \ip{\t}{\phi}$ by their parameters $\t$, and write $\phi(\tau) = \sum_{(s,a)\in\tau}\phi(s,a)$ for trajectory features.

Finally, we assume that the expert is optimal for $\tE$, with a known temperature. Appendix~\ref{apx:subopt_experts} extends our results to misspecified experts, e.g., when the assumed temperature is incorrect, in which case all our transferability results hold with an additional error proportional to the expert's suboptimality.




\section{Actively learning transferable rewards}

We now present our active method for learning transferable rewards minimizing~\eqref{eq:objective}, summarized in Algorithm~\ref{alg:methodology}. Its key idea is to query the expert under the most \emph{informative} dynamics, i.e., those revealing parts of the reward that matter for transfer across $\cP$ but remain poorly determined by previous demonstrations. The method proceeds in two phases. In a \emph{warm-up} phase, the learner has too little data to tell which dynamics are informative, so it queries the expert under each available dynamics in turn. It then switches to an \emph{active} phase, in which it uses its reward estimate to select the most informative dynamics.

To make this strategy precise, we first quantify the accuracy of reward estimates from adaptively collected demonstrations through ellipsoidal confidence bounds (Section~\ref{sec:conf}). These bounds determine when the estimate is reliable enough to guide dynamics selection. In Section~\ref{sec:dynamics_selection}, we use them to construct the two-phase selection algorithm, before deriving transfer guarantees for the learned reward in Section~\ref{sec:transfer}.

\begin{algorithm}[t]
\caption{Active IRL}
\label{alg:methodology}
\begin{algorithmic}[1]
\STATE \textbf{input:} set of dynamics $\cP$, warm-up length $n_0$
\STATE \textcolor{blue}{\texttt{// Warm-up phase}}
\FOR{round $k = 1,\dots,n_0$}
    \STATE Select $P_k$ round-robin over $\cP$~\eqref{eq:roundrobin}
    \STATE Observe $m$ trajectories in $P_k$
\ENDFOR
\STATE Estimate $\t_{n_0}$ by minimizing~\eqref{eq:minmaxirl}\label{line:irl_warmup}\COMMENT{Section~\ref{sec:conf}}
\STATE \textcolor{blue}{\texttt{// Active phase}}
\FOR{round $k = n_0+1,\dots,n$}
    \STATE Actively select $P_k$ via~\eqref{eq:active_selection_rule} \COMMENT{Section~\ref{sec:dynamics_selection}}
    \STATE Observe $m$ trajectories in $P_k$
    \STATE Estimate $\t_k$ by minimizing~\eqref{eq:minmaxirl}\label{line:irl}\COMMENT{Section~\ref{sec:conf}}
\ENDFOR
\end{algorithmic}
\end{algorithm}

\subsection{Reward estimation}
\label{sec:conf}
To select informative dynamics, we need to quantify how much information each dynamics provides about $\tE$, and how much has already been accumulated. We first introduce the loss and use its curvature to characterize the
information provided under each dynamics. We then show how the accumulated information yields an ellipsoidal confidence bound for $\tE$ (Lemma~\ref{thm:confidencenew}). Proofs are deferred to Appendix~\ref{apx:confidence_results}.

\textbf{IRL loss.}
After $k$ rounds, we estimate $\tE$ by minimizing the empirical loss over $\T$,
\begin{equation}
    \label{eq:minmaxirl}
    L_k(\t) \coloneq \sum_{i=1}^k l_i(\t),
    \quad
    l_i(\t) \coloneq V^\star_{P_i}\br{\Pi_{P_i}\t} - \ip{\t}{\Pi_{P_i}\hat\phi_i},
\end{equation}
and write $\t_k \in \argmin_{\t\in\T}L_k(\t)$. Here $P_i$ are the dynamics selected at round $i$, $\hat\phi_i = \frac1m\sum_{j=1}^m\phi(\tau_{i,j})$ is the empirical mean of the expert's trajectory features, and $\Pi_{P_i}$ is the orthogonal projector onto the identifiable subspace $U_{P_i}$ defined below. Without the projections, each $l_i$ is an entropy-regularized version of the min-max IRL loss~\citep{syed2007game} under $P_i$.

\textbf{Curvature and identifiability.}
The optimal policy induced by a reward depends on the underlying dynamics. Hence, by selecting the dynamics under which the expert demonstrates, the learner chooses which parts of the reward can be estimated, and how efficiently. We characterize this through the Hessian of the optimal value function,
\[
    \cH_P(\t) \coloneq \nabla_\t^2 V^\star_P(\t),
\]
which is the \emph{information} provided by $P$. It is also the curvature of the per-round loss $l_i$ when $P = P_i$, since it differs 
from $V^\star_{P_i}$ only by a linear term. For a direction $v\in\R^d$, 
the curvature $\norm{v}_{\cH_P(\t)}^2$ quantifies the sensitivity of the optimal policy $\pi_P(\t)$ to reward perturbations along $v$. The larger the curvature, the fewer demonstrations under $P$ are needed to estimate the reward along $v$. With zero curvature, the reward along $v$ cannot be estimated under $P$, regardless of how many trajectories are collected. Such directions arise from potential shaping~\citep{ng1999policy} and from reward changes at unreachable states. We therefore define the \emph{identifiable subspace} $U_P\coloneq\im\cH_P(\t)$, the orthogonal complement of the zero-curvature directions.\footnote{$U_P$ does not depend on $\t$, see Appendix~\ref{apx:confidence_results}.} Demonstrations under $P$ only inform the components of $\tE$ in $U_P$. The projections in~\eqref{eq:minmaxirl} thus prevent sampling noise in $\hat\phi_i$ from drifting the estimate along unidentifiable directions (Appendix~\ref{apx:confidence_results}). Across all available dynamics, the identifiable components of $\tE$ lie in the joint identifiable subspace $U\coloneq\sum_{P\in\cP}U_P$, of dimension $d_U\le d$.

\textbf{Confidence sequence.}
After $k$ rounds, the information accumulated about $\tE$ is given by the curvature of $L_k$~\eqref{eq:minmaxirl} at $\tE$ and its smallest eigenvalue on $U$,
\[
    \cH_k \coloneq \sum_{i=1}^k\cH_{P_i}(\tE),
    \qquad
    \lambda_k \coloneq \min_{\substack{v \in U\\ \;\norm{v}_2=1}}\norm{v}^2_{\cH_k},
\]
where $\lambda_k$ measures the information in the least-informed identifiable direction. Our main estimation result is a confidence sequence for $\tE$, whose radius depends on the effective dimension
\begin{equation}\label{eq:dstar}
    d_\star \coloneq d_U \max_{P\in\cP}
    \norm{\cH_P(\tE)^{\dagger/2}\,\Sigma_P\,\cH_P(\tE)^{\dagger/2}}_2,
\end{equation}
where $\Sigma_P \coloneq \cov_P^{\piE_P}\br{\Pi_P\phi(\tau)}$ is the covariance of the projected expert trajectory features.
\begin{restatable}{lemma}{confidencenew}
\label{thm:confidencenew}
Fix $n\in\mathbb N^+$, $\delta\in(0,1)$, and $R\coloneq\eta B_\phi H(H+1)$. Let $n_\delta$ be the first round $k\le n$ such that
\begin{equation}\label{eq:n_min}
    \lambda_k \;\ge\; \frac{32R^2d_\star}{(1-e^{-1})^2m}\log\frac{3\eta d_\star nm}{\delta},
\end{equation}
with $n_\delta=\infty$ if no such round exists. Then, with probability at least $1-\delta$, simultaneously for all $n_\delta\le k\le n$,
\[
    \norm{\t_k-\tE}_{\cH_k}^2 \;\le\; \frac{16d_\star}{(1-e^{-1})^2m} \log\frac{3\eta d_\star km}{\delta} \;\eqqcolon\; \beta_k^2(\delta).
\]
\end{restatable}
\begin{proof}[Proof sketch]
\vspace{-0.25cm}
We combine concentration of the projected empirical features under adaptive selection with the pseudo self-concordance of $L_k$~\citep{bach2010self}. After the warm-up, the localization approach of~\citet{schlaginhaufen2026fastratesinversereinforcement} converts sampling variation into an ellipsoidal parameter-error bound.
\vspace{-0.25cm}
\end{proof}
Lemma~\ref{thm:confidencenew} quantifies the uncertainty in our reward estimate through a confidence ellipsoid shaped by the accumulated information $\cH_k$. The ellipsoid is wide along directions where little information has been collected, and narrow where much has. Querying the expert under $P$ adds $\cH_P(\tE)$ to the accumulated information, and is therefore informative when $P$ provides information along directions where the ellipsoid is wide.

The ellipsoid, however, is only guaranteed to contain $\tE$ once enough information has been accumulated in every identifiable direction, as prescribed by the warm-up condition~\eqref{eq:n_min}.  Since $L_k$ is nonquadratic, its curvature changes with the reward parameters at a rate controlled by $R$~\citep{bach2010self}. The warm-up ensures that $\t_k$ lies in a region where the curvature remains comparable to that at $\tE$, allowing the approximately local quadratic analysis that yields the confidence bound~\citep{chaudhuri2015convergence,lee2025improved}.

The effective dimension $d_\star$ compares the variability of the expert's features to the information they carry, and plays the role of the parameter dimension in the confidence radius. It satisfies $d_\star\ge d_U/\eta$, with equality under deterministic dynamics, where the projections also have no effect on the loss~\eqref{eq:minmaxirl}. Together with the curvature variation controlled by $R$, it determines the required information level~\eqref{eq:n_min}. Through $R^2$, the threshold has an explicit quartic dependence on the horizon $H$ and quadratic dependence on the feature scale $B_\phi$. Up to logarithmic factors, both the threshold and the squared confidence radius $\beta_k^2(\delta)$ scale as $1/m$. Larger batches therefore relax the warm-up requirement and shrink the confidence ellipsoid by reducing the sampling noise in the empirical features.

Lemma~\ref{thm:confidencenew} thus splits dynamics selection into two goals: first accumulating enough information in every direction of $U$ to complete the warm-up, then adding information where it matters for transfer across $\cP$, which we address next.

\subsection{Dynamics selection}
\label{sec:dynamics_selection}
We now use the confidence sequence to select dynamics so that the learned reward transfers across $\cP$~\eqref{eq:objective}. We bound the suboptimality under each $P\in\cP$, then derive a two-phase algorithm querying the dynamics where this bound is worst. Proofs are in Appendix~\ref{apx:dynamics_selection}.

\textbf{From information to suboptimality.}
Lemma~\ref{thm:confidencenew} bounds the error $\t_k-\tE$ in the geometry of the accumulated information $\cH_k$. Transfer to $P$, however, is governed by the geometry of $\cH_P(\tE)$, which measures how sensitive the optimal policy under $P$ is to reward errors (Section~\ref{sec:conf}). We compare the two through the \emph{information gain} of $P$,
\begin{equation}
\label{eq:transfer_coefficient}
    g_k(P) \coloneqq \norm{\cH_k^{\dagger/2}\,\cH_P(\tE)\,\cH_k^{\dagger/2}}_2,
\end{equation}
which is large when $P$ provides much information along directions where little has been accumulated. Errors along such directions remain large, and since they are important for transfer to $P$, they are amplified in the suboptimality under $P$.
\begin{restatable}{lemma}{transferability}
\label{lemma:transfer}
Fix $n\in \mathbb{N}^+$ and $\delta\in(0,1)$. With probability at least $1-\delta$, it holds for all $P \in \cP$ and $n_\delta \le k \le n$ that
\[
\subopt(\t_k,P)\;\le\; g_k(P)\,\beta^2_k(\delta).
\]
\end{restatable}
\begin{proof}[Proof sketch]
Let $u_k \coloneq \t_k - \tE$. The suboptimality is the Bregman divergence of $V^\star_P$ between $\tE$ and $\t_k$, i.e., $\subopt(\t_k,P) = \int_0^1 (1-t)\,\norm{u_k}^2_{\cH_P(\t_k - t u_k)}\,\mathrm{d}t$. On the event of Lemma~\ref{thm:confidencenew}, for every $P\in\cP$ and $n_\delta\le k\le n$,
\vspace{-0.3cm}
\[
    \subopt(\t_k,P)
    \;\overset{\text{(i)}}{\lesssim}\; \norm{u_k}^2_{\cH_P(\tE)}
    \;\overset{\text{(ii)}}{\le}\; g_k(P)\,\norm{u_k}^2_{\cH_k}.
\]
(i) holds by pseudo self-concordance of $V^\star_P$, since $\t_k$ is close to $\tE$ after the warm-up. (ii) changes the geometry from $\cH_P(\tE)$ to $\cH_k$ at a cost $g_k(P)$. Lemma~\ref{thm:confidencenew} bounds the last term by $\beta_k^2(\delta)$.
\vspace{-0.25cm}
\end{proof}
Lemma~\ref{lemma:transfer} suggests that transfer is hardest to the dynamics with the most information left to gain, which are thus the most valuable to query. Since $\beta_k^2(\delta)$ does not depend on dynamics, our goal is to make $g_k(P)$ small across $\cP$. However, $g_k(P)$ depends on the unknown $\tE$, and plugging in $\t_k$ is only reliable after the warm-up, so we first collect data passively.

\textbf{Warm-up.}
\label{sec:warmup}
During the warm-up, we cycle through the available dynamics using a \emph{round-robin} rule that does not rely on reward estimates. Writing $k(P)$ for the number of times $P$ was selected in the first $k$ rounds, we choose the next dynamics according to
\begin{equation}
    P_{k+1} \in \argmin_{P\in \cP} k(P).
    \label{eq:roundrobin}
\end{equation}
This simple rule guarantees that the warm-up condition is met within a bounded number of rounds.
\begin{restatable}{lemma}{roundrobin}
\label{lemma:passive_learning}
Fix $n\in \mathbb{N}^+$ and $\delta \in (0,1)$. Under round-robin selection~\eqref{eq:roundrobin}, the warm-up length satisfies
\[
    n_\delta \;\le\; \abs{\cP}\roof{\frac{2R^2\beta_n^2(\delta)}{\lambda_{\abs{\cP}}}},
\]
whenever the right-hand side is at most $n$.
\vspace{-0.25cm}
\end{restatable}
\begin{proof}[Proof sketch]
After $k$ rounds, every $P\in\cP$ was picked at least $\floor{k/\abs{\cP}}$ times, so $\lambda_k \ge \floor{k/\abs{\cP}}\,\lambda_{\abs{\cP}}$.
\vspace{-0.25cm}
\end{proof}
The threshold~\eqref{eq:n_min} equals $2R^2\beta_n^2(\delta)$, and $\lambda_{\abs{\cP}}$ is the information in the least-informed direction after one sweep over $\cP$. The bound thus counts the sweeps needed before Lemma~\ref{thm:confidencenew} holds. It depends on the unknown $\tE$, although a loose $\tE$-independent version can be computed (Appendix~\ref{app:computable_warmup}). A similar dependence appears in active learning for maximum-likelihood estimation~\citep{chaudhuri2015convergence}, where a passive phase provides an initial estimate of the Fisher information used to select informative labels. Here, the warm-up makes the estimated Hessians reliable enough to guide dynamics selection. Being worst-case, the bound is conservative, and Section~\ref{sec:experiments} shows that the active phase performs well after a much shorter warm-up.

\textbf{Active selection.}
After the warm-up, $\t_k$ is localized around $\tE$ (Lemma~\ref{thm:confidencenew}). This is not yet enough to transfer well to dynamics that rely on weakly informed directions, but it is enough to estimate information gains, since the Hessians at $\t_k$ then match those at $\tE$ up to a constant factor. We thus estimate $g_k(P)$ by $\hat g_k(P)$, defined as in~\eqref{eq:transfer_coefficient} with Hessians at $\t_k$ instead of $\tE$, and query the dynamics with the largest estimated information gain,
\begin{equation}
\label{eq:active_selection_rule}
    P_{k+1} \in \argmax_{P \in \cP}\; \hat g_k(P).
\end{equation}
Querying the selected dynamics adds its information to $\cH_k$, lowering its information gain as well as that of every dynamics sharing its identifiable directions. Section~\ref{sec:transfer} shows that this rule drives the worst-case suboptimality down across $\cP$.

\begin{remark}
\label{rem:gopt}
Minimizing $\max_{P\in\cP} g_k(P)$ relates to G-optimal designs~\citep{kiefer1959optimum,kiefer1974general}. If each $\cH_P(\t)$ were rank-one and parameter-independent, our rule~\eqref{eq:active_selection_rule} would reduce to greedy D-optimal design. More generally, Theorem~\ref{theo:active_learning} still holds if the spectral norm in the selection rule~\eqref{eq:active_selection_rule} is replaced by the trace or by $\log\det(I+\cdot)$. Our experiments use the latter.
\end{remark}
\begin{remark}[Offline active learning]
\label{rem:offline}
When sequential interaction is impractical, one can select all dynamics in advance by using $\t_0=0$ in place of $\t_k$ (Appendix~\ref{app:offline_active_learning}).
\end{remark}

\textbf{Computational complexity.}
The information $\cH_P(\t) = \eta\,\EX\big[A^\t_P(\tau)A^\t_P(\tau)^\top\big]$ is the second moment of the feature advantages $A^\t_P(\tau)$ along trajectories of $\pi_P(\t)$ (Appendix~\ref{apx:confidence_results}). It is estimated by sampling such trajectories and computing advantages through one policy evaluation.
Since $U_P$ does not depend on $\t$, each projector $\Pi_P = \cH_P(0)\cH_P(0)^\dagger$ is computable once at the start, using the uniform policy $\pi_P(0)$ (Appendix~\ref{app:algorithms}). As in any IRL method, the cost of Algorithm~\ref{alg:methodology} is dominated by the RL solves in lines~6 and~11. Active selection only adds line~9, which reuses the policies computed by IRL to estimate the information, whereas \citet{buening2024environment} require $\cO(k\abs\cP)$ RL solves at round $k$ to choose dynamics.

\subsection{Transfer guarantees}
\label{sec:transfer}
We now show that our two-phase approach yields a reward that transfers across $\cP$ at a rate independent of $\abs\cP$, and extend this guarantee to dynamics outside $\cP$. Proofs of this section are deferred to Appendix~\ref{apx:transfer}.

\textbf{Transfer across $\cP$.}
The following theorem shows that, as active rounds accumulate, less and less information remains to be gained under any $P\in\cP$, and the worst-case suboptimality decreases accordingly.
\begin{restatable}{theorem}{greedy}
\label{theo:active_learning}
Fix $n\in \mathbb{N}^+$, $\delta \in (0,1)$, and follow the active rule~\eqref{eq:active_selection_rule} after round $n_\delta$. With probability at least $1-\delta$, it holds for all $n_\delta<k\le n$ and $P\in\cP$ that
\[
    g_k(P) \;\lesssim\; \br{1+\frac{m}{\eta d_\star H^2}}\frac{d_U\log(km)}{k-n_\delta}.
\]
If moreover $m\le\eta d_\star H^2$ and $n\ge 2n_\delta$,
\begin{equation}
\label{eq:main_rate}
    \subopt(\t_n,P) \;\lesssim\; \frac{d_\star\,d_U\log(nm)\log(\eta d_\star nm/\delta)}{n\,m}.
\end{equation}
\end{restatable}
\begin{proof}[Proof sketch]
For all $P\in\cP$, $\cH_P(\t_k)$ and $\cH_P(\tE)$ are within an $e$ factor of each other. An elliptical potential argument relating the spectral norm to the log-determinant bounds $g_k(P)$ for all $P \in \cP$. For $m\le\eta d_\star H^2$, the front factor is at most $2$, and $n\ge 2n_\delta$ gives $n-n_\delta\ge n/2$. Lemma~\ref{lemma:transfer} then yields~\eqref{eq:main_rate}.
\vspace{-0.25cm}
\end{proof}
Theorem~\ref{theo:active_learning} shows that, after the warm-up, the information gain of every $P\in\cP$ decays inversely with the number of active rounds $k-n_\delta$, so that progressively no dynamics in $\cP$ is left with much to reveal. With at most $\eta d_\star H^2$ trajectories per round, and as long as the warm-up uses at most half the budget, the returned reward thus achieves a worst-case suboptimality of $\tilde\cO\br{d_\star d_U/(nm)}$ over $\cP$, i.e., inversely proportional to the total number of trajectories.

Importantly, $\abs\cP$ only affects the warm-up length (Lemma~\ref{lemma:passive_learning}), not the suboptimality rate. Round-robin instead pays for $\abs\cP$ at every round. If only one $P\in\cP$ is informative and the others are redundant, our analysis only guarantees a suboptimality of $\tilde\cO\br{d_\star\abs\cP/(nm)}$ for round-robin, as it keeps querying dynamics with little information to offer. Active selection thus replaces the number of dynamics $\abs\cP$ by the dimension $d_U$, an improvement whenever $\abs\cP > d_U$.

Intuitively, the condition $m\le\eta d_\star H^2$ states that when reconfiguring the dynamics is costly, the expert can demonstrate up to $\eta d_\star H^2$ trajectories per round without loss. Beyond this threshold, the noise in the empirical features $\hat\phi_i$ in~\eqref{eq:minmaxirl} still shrinks, but the information each round contributes is set by its dynamics and does not grow with $m$, so our bound no longer improves.

\textbf{Transfer beyond $\cP$.}
We now extend our guarantee to target dynamics $P$ outside $\cP$. Such dynamics may require information along directions that no dynamics in $\cP$ identifies, which we measure by $g_\perp(P) \coloneqq \norm{\Pi_{U^\perp}\cH_P(\tE)\Pi_{U^\perp}}_2$, where $U^\perp$ is the orthogonal complement of the identifiable subspace $U$.

\begin{restatable}{theorem}{transferabilitybeyond}
\label{theorem:transfer}
Fix $\delta \in (0,1)$  and follow the active rule~\eqref{eq:active_selection_rule} after round $n_\delta$. If $m\le\eta d_\star H^2$ and $n\ge 2n_\delta$, then with probability at least $1-\delta$, for any $P$,
\[
    \subopt(\t_n,P) \;=\; \tilde\cO\br{\frac{d_\star\,d_U}{n\,m}} \;+\; \cO\br{g_\perp(P)\,B_\t^2}.
\]
\end{restatable}
\begin{proof}[Proof sketch]
Let $\t_n-\tE=u+v$ with $u\in U$ and $v\in U^\perp$. Theorem~\ref{theo:active_learning} bounds the error along $u$, up to how much information $\cP$ provides along the directions that matter under $P$. Since $\norm{v}_2\le 2B_\t$, we have $\norm{v}_{\cH_P(\tE)}^2\le 4g_\perp(P)B_\t^2$. Pseudo self-concordance then yields the result, stated in full in Appendix~\ref{apx:transfer}.
\vspace{-0.25cm}
\end{proof}
Theorem~\ref{theorem:transfer} splits suboptimality under \emph{any} $P$ into two terms. The first shrinks with the number of demonstrations, and corresponds to directions shared by $U_P$ and $U$. The second is a transfer bias, and corresponds to directions that no dynamics in $\cP$ cover. It therefore cannot be reduced by observing the expert in $\cP$. Interestingly, our active rule helps reduce the first term, but only the choice of $\cP$ can mitigate the second. When $U_P\subseteq U$, the bias term vanishes, and in particular for $P\in\cP$, we exactly recover~\eqref{eq:main_rate}.

\section{Experiments}\label{sec:experiments}
Our experiments serve two purposes: \emph{(i)} to evaluate when active learning improves sample efficiency in terms of expert demonstrations, and \emph{(ii)} to identify which dynamics are most useful for efficiently learning transferable rewards.

We instantiate our two-phase strategy (Section~\ref{sec:dynamics_selection}) with the log-determinant criterion (Remark~\ref{rem:gopt}), which performed best among valid criteria we compare in Appendix~\ref{app:additional_experiment_details}. Since $n_\delta$ cannot be computed, we treat the warm-up length as a hyperparameter, set to one third of the budget, and study other choices in Appendix~\ref{app:additional_experiment_details}. We compare against baselines that isolate the benefits of \emph{(a)} the warm-up phase, \emph{(b)} leveraging multiple dynamics, and \emph{(c)} learning transferable rewards rather than directly imitating behavior.

To evaluate \emph{(a)}, we run an \emph{Active only} ablation that follows the active rule~\eqref{eq:active_selection_rule} from the start.\footnote{This uses a regularized accumulated Hessian $\cH_k+\varepsilon \Pi_U$ to make $\hat g_k(P)$ well defined.} To evaluate \emph{(b)}, we compare against the environment design (ED) approach of~\citet{buening2024environment}, a round-robin baseline following~\eqref{eq:roundrobin}, and methods that learn from one fixed dynamics in $\cP$: \emph{Fixed dynamics} IRL and GAIL~\citep{ho2016generative}.  All IRL methods except GAIL optimize the min-max IRL objective~\eqref{eq:minmaxirl}. Finally, to evaluate \emph{(c)}, we compare against imitation methods: Behavioral Cloning (BC), trained on a single dynamics like the methods above, and robust imitation via RIME~\citep{chae2022robust}.

All methods follow the same evaluation protocol. At each round $k$, the learner selects dynamics $P_k \in \cP$ and observes $m=1$ expert trajectory, generated by rolling out, under $P_k$, the entropy-regularized optimal policy for the true reward parameters $\tE$. For fixed-dynamics methods, the environment is uniformly sampled once from $\cP$ and then fixed. Evaluation proceeds by learning a reward from collected data and deploying a policy optimized for each tested dynamics; imitation methods (BC, RIME) instead deploy a single policy across all dynamics. In all experiments, state–action features are one-hot encodings over states and actions.

We measure performance by the worst-case suboptimality~\eqref{eq:objective} and by the reward error up to constants, $\min_{c\in \mathbb{R}} \norm{\tE - \t_n + c\mathbf{1}}_2$.  We report means and standard errors over 20 randomly seeded runs. These experiments focus on tractable environments that require no approximation, leaving scalability to future work. 

\subsection{Controlled sets of dynamics}
 \label{subsec:controlled_dynamics}

\begin{figure*}[t]
    \centering
    \includegraphics[width=0.85\textwidth]{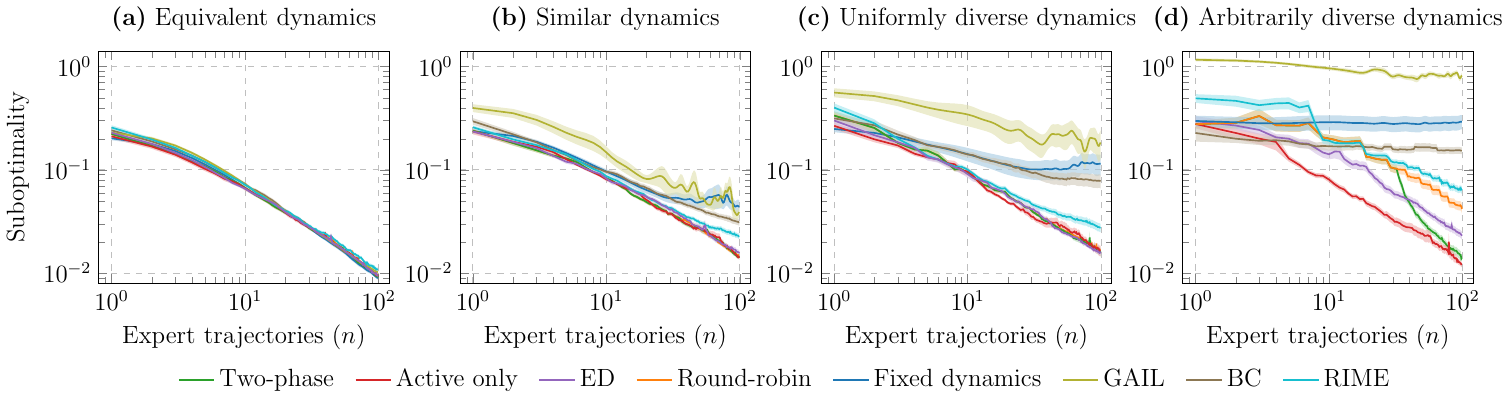}
\caption{
Worst-case suboptimality over the training set $\cP$ as a function of the number of observed expert trajectories on randomly generated MDPs. Each plot corresponds to a different type of training set.
\vspace{-0.5em}
}
    \label{fig:angle_curves}
\end{figure*}
\begin{figure}[h!]
    \centering
    \includegraphics[width=\columnwidth]{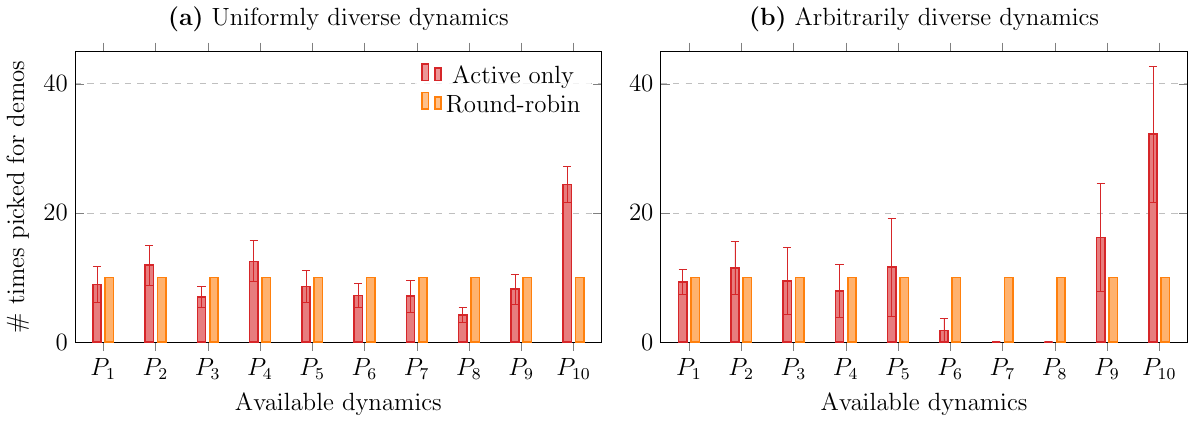}
    \caption{Number of times each dynamics in the training set $\cP$ is selected after 100 rounds, for two types of training set. 
    Dynamics are ordered by increasing average dissimilarity to the rest of $\cP$, so that $P_{10}$ corresponds to the most isolated dynamics.
    }
\label{fig:angles_picked}
\vspace{-1.5em}
\end{figure}

We construct training sets $\cP$ in which we control how well rewards learned in one dynamics transfer to the others. This is quantified by principal angles between unidentifiable subspaces~\citep{Galantai2004, schlaginhaufen2024towards} (see also Appendix~\ref{section:principal}). 

We consider four types of sets:
(1) Equivalent dynamics, where rewards transfer perfectly across $\cP$; (2) Similar dynamics, where rewards transfer consistently well; (3) Uniformly diverse dynamics, where transfer is uniformly weak across pairs; (4) Arbitrarily diverse dynamics, where transferability is heterogeneous across the set. For each random seed, we generate a set of $\abs{\cP}=10$ dynamics. After $n=100$ demonstrations, methods are also evaluated on unseen dynamics.
The MDPs all have $|\mathcal{S}|=20$ states and $|\mathcal{A}|=5$ actions.

\textbf{Results.}
With equivalent dynamics (Figure~\ref{fig:angle_curves}(a)), all methods perform similarly. Since every dynamics reveals the same information about the reward, dynamics selection offers no advantage, and imitation methods remain competitive with reward learning. As dynamics become more diverse (Figures~\ref{fig:angle_curves}(b)--(d)), fixed-dynamics methods (IRL, GAIL, and BC) increasingly stall. For reward learning, this is the bias of Theorem~\ref{theorem:transfer} with a single training dynamics, which grows with the diversity of $\cP$. Active selection is most beneficial in arbitrarily diverse sets (Figure~\ref{fig:angle_curves}(d)). There, the Active only variant performs best, both on $\cP$ and on unseen dynamics, where it substantially outperforms ED (Table~\ref{tab:transferability}). This suggests that our warm-up length is conservative, a gap we discuss in Section~\ref{sec:conclusion}. 

Figure~\ref{fig:angles_picked} shows which dynamics are selected. In uniformly diverse sets, selection closely matches round-robin. In arbitrarily diverse sets, it concentrates on dynamics that either transfer well on average across $\cP$ ($P_1,\dots,P_5$) or are highly isolated from the rest ($P_{10}$).

\subsection{Random mazes}
\begin{figure*}[t!]
    \centering
    \includegraphics[width=0.69\textwidth]{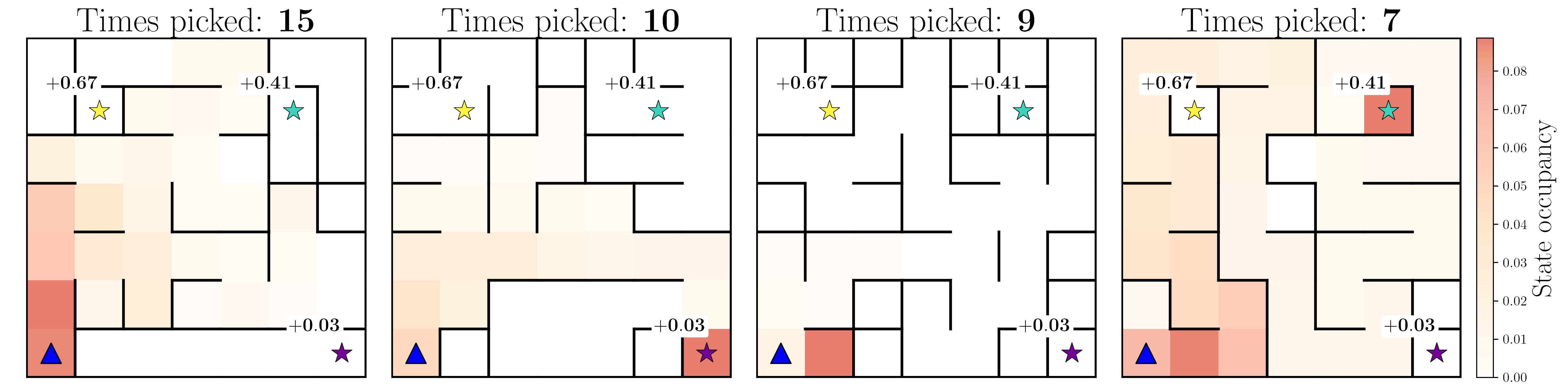}
\caption{Mazes most frequently selected by Active only after 100 rounds, on a representative seed. The blue triangle marks the initial state and stars mark goal states, labeled with their expert rewards. The heatmap shows the state occupancy of policies optimized for the inferred reward.}
\label{fig:maze_picked}
\vspace{-1em}
\end{figure*}

We next consider a maze navigation task~\citep{buening2024environment}, where the agent is rewarded for reaching and waiting in one of three goal states. This task makes the selected dynamics easy to visualize (Figure~\ref{fig:maze_picked}). We generate sets of $\abs\cP=100$ mazes by randomly placing walls, ranging from nearly empty layouts to highly constrained or even unsolvable ones (e.g., blocked goals). All mazes in a set share the initial state and the reward, with goal rewards sampled uniformly in $[0,1]$. Here, $|\cS| = 49$ and $|\cA| = 5$ (four moves and a waiting action).

\begin{figure}[h!]
    \centering
    \includegraphics[width=\columnwidth]{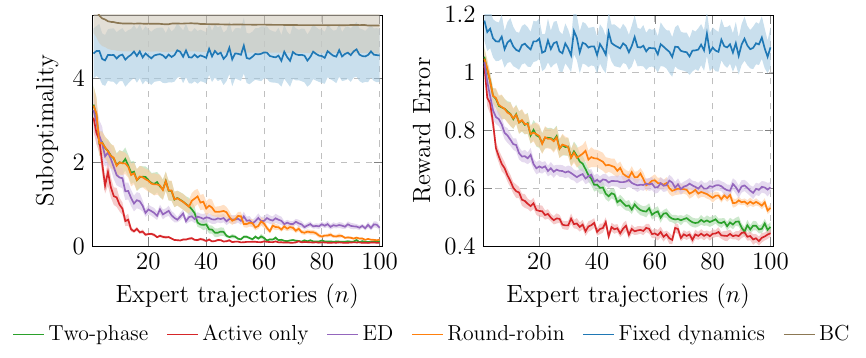}
    \caption{Worst-case suboptimality over the training set $\cP$ of $100$ random mazes, and reward error, as a function of the number of observed expert trajectories.}
    \label{fig:maze_curves}
    \vspace{-1em}
\end{figure}

\textbf{Results.}
Active selection is clearly more sample-efficient (Figure~\ref{fig:maze_curves}). Without warm-up, it reaches with 20 trajectories the suboptimality that round-robin attains with 100. The two-phase strategy catches up after about 60 trajectories, again suggesting that the warm-up is conservative. 

Figure~\ref{fig:maze_picked} shows that the active rule often selects unsolvable mazes. When goals are all unreachable but the layout remains open, the entropy-regularized expert acts uniformly, revealing that the reward is constant across reachable non-goal states. A maze can thus be informative without being solvable.

\begin{table}[t]
\centering
\small
\caption{Worst-case suboptimality ($\times 100$) over 1000 \emph{unseen}  dynamics after learning from $n=100$ expert trajectories, collected on either a uniformly or an arbitrarily diverse training set of dynamics.}
\label{tab:transferability}
\setlength{\tabcolsep}{6pt}
\begin{tabular}{lcc}
\toprule
& \multicolumn{2}{c}{Training set type} \\
\cmidrule(lr){2-3}
\textbf{Algorithm} & \textbf{Uniform} & \textbf{Arbitrary} \\
\midrule
Active only    & $\mathbf{\phantom{0}2.86 \pm 0.02}$ & $\mathbf{12.63 \pm 0.14}$ \\
Two-phase      & $\phantom{0}2.93 \pm 0.02$ & $13.63 \pm 0.15$ \\
\hdashline
ED             & $\phantom{0}3.32 \pm 0.02$ & $16.52 \pm 0.22$ \\
Round-robin   & $\phantom{0}3.15 \pm 0.02$ & $19.59 \pm 0.19$ \\
Fixed dynamics & $18.18 \pm 0.16$ & $24.36 \pm 0.09$ \\
GAIL           & $26.62 \pm 0.37$ & $96.06 \pm 0.42$ \\
BC             & $15.05 \pm 0.18$ & $20.01 \pm 0.10$ \\
RIME           & $\phantom{0}4.91 \pm 0.03$ & $24.47 \pm 0.34$ \\
\bottomrule
\end{tabular}
\vspace{-1em}
\end{table}



\section{Related Work}

\textbf{Active IRL} aims to minimize expert interactions, the main practical bottleneck, by querying the expert selectively. Prior work has mostly focused on fixed dynamics. \citet{Lopes2009ActiveLFA} and \citet{Cui2018ActiveRL} select states or trajectories that maximize information about the reward, \citet{bajgar2025pac} provide PAC guarantees for such Bayesian state queries, and \citet{lindner2023activeexplorationinversereinforcement} use confidence sets over feasible rewards to guide exploration. Our setting instead queries the expert under chosen dynamics.

\textbf{Learning from given dynamics.} In the entropy-regularized setting, sufficiently diverse dynamics allow recovering rewards up to constants when the expert policy is known~\citep{cao2021identifiability,rolland2022identifiability}. For transfer from demonstrations, \citet{schlaginhaufen2024towards} give sufficient conditions, and \citet{hao2026reward} derive finite-sample guarantees from one source to a given target environment. These results require the dynamics to be well covered, through ergodicity~\citep{cao2021identifiability,rolland2022identifiability,schlaginhaufen2024towards} or through source data covering the target's occupancy~\citep{hao2026reward}, whereas none of our results require either assumption. Our confidence sets extend the fast rates of \citet{schlaginhaufen2026fastratesinversereinforcement} from fixed dynamics to adaptively selected ones. Without regularization, rewards are only partially identifiable, and recent work selects rewards that remain robust to this ambiguity~\citep{lazzati2026robustness,lazzati2025generalizing}. Beyond reward learning, \citet{chae2022robust} learn a single policy robust to dynamics variations by imitating experts on randomly sampled dynamics. None of these works lets the learner choose the dynamics in which to collect demonstrations.

\textbf{Choosing dynamics.} \citet{amin2017repeated} show that an agent choosing the tasks can identify the expert's reward when the expert policy is revealed in each, and bound the number of mistakes when tasks are chosen adversarially. \citet{nitschke2024amber} maximize behavior entropy, \citet{Bajgar2024TowardITA} maximize expected information gain, and \citet{buening2024environment} target where the current reward estimate is expected to  perform worst. These methods focus on state-only rewards or deterministic experts, rely on expensive selection rules, or lack sample complexity guarantees for dynamics selection. In contrast, our work provides a practical and provably efficient framework for actively choosing dynamics to learn transferable rewards.

\section{Conclusion}
\label{sec:conclusion}
We introduced an active learning method for inferring transferable rewards, which adaptively selects the dynamics under which the expert demonstrates. After a round-robin warm-up, it queries the expert under the dynamics with the largest information gain. The resulting reward transfers across all available dynamics at a rate governed by the dimension of the identifiable reward subspace rather than their number, and to unseen dynamics up to a bias from the directions left uncovered. Empirically, it requires substantially fewer demonstrations than passive selection and outperforms the most closely related active approach~\citep{buening2024environment}.

\textbf{Limitations and future work.}
First, our guarantees rely on a conservative warm-up, while active selection performs well without it in practice, possibly because it naturally first selects the dynamics informing directions not yet covered. Additional assumptions on the information gathered, ensuring that our local curvature argument applies early, or data-driven switching criteria could close this gap. Second, IRL across multiple dynamics requires one optimal policy per dynamics, which is costly for large $\abs\cP$, and dynamics-conditioned policies could amortize this cost. Finally, our study is restricted to finite sets of tabular environments and linear rewards. A promising direction is to model rewards with neural networks whose last layer is linear, using the last hidden layer as learned low-dimensional features $\phi$ and computing Hessians with respect to the last layer only, as in~\citet{ash2021gone}.


\bibliography{references}

\section*{Checklist}



\begin{enumerate}

  \item For all models and algorithms presented, check if you include:
  \begin{enumerate}
    \item A clear description of the mathematical setting, assumptions, algorithm, and/or model. Yes. Section~\ref{sec:formulation} for the setting, assumptions and models. Section~\ref{sec:dynamics_selection} for algorithms.
    \item An analysis of the properties and complexity (time, space, sample size) of any algorithm. Yes. Sample complexity is analyzed in Lemma~\ref{lemma:passive_learning} and Theorem~\ref{theo:active_learning}, and computational complexity in the "Computational complexity" paragraph of Section~\ref{sec:dynamics_selection} and Appendix~\ref{app:algorithms}.
    \item (Optional) Anonymized source code, with specification of all dependencies, including external libraries. Yes. Link to the anonymized source code provided in Appendix~\ref{app:additional_experiment_details}.
  \end{enumerate}

  \item For any theoretical claim, check if you include:
  \begin{enumerate}
    \item Statements of the full set of assumptions of all theoretical results. Yes. See Section~\ref{sec:formulation}.
    \item Complete proofs of all theoretical results. Yes, all proofs are provided in Appendices~\ref{apx:confidence_results} and~\ref{apx:dynamics_selection}.
    \item Clear explanations of any assumptions. Yes.
  \end{enumerate}

  \item For all figures and tables that present empirical results, check if you include:
  \begin{enumerate}
    \item The code, data, and instructions needed to reproduce the main experimental results (either in the supplemental material or as a URL). Yes. We provide a link to the code used in experiments in Appendix~\ref{app:additional_experiment_details}.
    \item All the training details (e.g., data splits, hyperparameters, how they were chosen). Yes, hyperparameters are specified both in Appendix~\ref{app:additional_experiment_details}, and through the opensourced code used for our experiments.
    \item A clear definition of the specific measure or statistics and error bars (e.g., with respect to the random seed after running experiments multiple times). Yes. We consistently provide the number of seeds used, and what error bars represent in both the main text (Section~\ref{sec:experiments}) and appendix.
    \item A description of the computing infrastructure used. (e.g., type of GPUs, internal cluster, or cloud provider). Yes. Information is included in Appendix~\ref{app:additional_experiment_details}.
  \end{enumerate}

  \item If you are using existing assets (e.g., code, data, models) or curating/releasing new assets, check if you include:
  \begin{enumerate}
    \item Citations of the creator If your work uses existing assets. Not Applicable.
    \item The license information of the assets, if applicable. Not Applicable.
    \item New assets either in the supplemental material or as a URL, if applicable. Not applicable.
    \item Information about consent from data providers/curators. Not Applicable. 
    \item Discussion of sensible content if applicable, e.g., personally identifiable information or offensive content. Not Applicable.
  \end{enumerate}

  \item If you used crowdsourcing or conducted research with human subjects, check if you include:
  \begin{enumerate}
    \item The full text of instructions given to participants and screenshots. Not Applicable.
    \item Descriptions of potential participant risks, with links to Institutional Review Board (IRB) approvals if applicable. Not Applicable.
    \item The estimated hourly wage paid to participants and the total amount spent on participant compensation. Not Applicable.
  \end{enumerate}

\end{enumerate}

\clearpage
\appendix
\thispagestyle{empty}

\onecolumn
\aistatstitle{Appendix}

\section{Notations and Preliminaries} 
\label{app:notation}
\subsection{Notations}
We enumerate below the most important notations used in our appendix.
\begin{table}[th!]
\centering
\label{table_symbols}
\small
\begin{tabular}{lll}
\toprule
\multicolumn{3}{l}{\emph{Problem setting}}\\
\midrule
$\cP$ & $\coloneq$ & finite set of dynamics available for expert demonstrations\\
$n,\ m$ & $\coloneq$& number of rounds, and expert trajectories per round\\
$\tE,\ \T$ &$\coloneq$ & expert reward parameters, and $\T\coloneq\bc{\t \in \R^d: \norm{\t}_2 \le B_\t}$\\
$\pi_P(\t)$ & $\coloneq$ & $\arg\max_{\pi} V_{P}^\pi(\t)$, optimal regularized policy\\
$\nu^{\pi}_{h,P}(s)$ & $\coloneq$ & $\mathbb{P}_P^{\pi}\bs{s_h=s}$\\
$\mu^{\pi}_{h,P}(s,a)$ & $\coloneq$ & $\nu^\pi_{h,P}(s)\,\pi_h(a \mid s)$\\
$\operatorname{Ent}_h(\pi)$ & $\coloneq$ & $-\sum_{s,a}\mu^\pi_{h,P}(s,a)\log\pi_h(a \mid s)$\\
$\phi(\tau)$ & $\coloneq$ & $\sum_{h=1}^H \phi(s_h, a_h)$\\
$\phi_P(\t),\ \phi^E_P$ & $\coloneq$ & $\mathbb{E}_P^{\pi_P(\t)}\bs{\phi(\tau)}$, and $\phi^E_P \coloneq \phi_P(\tE)$\\
$\subopt(\t,P)$ & $\coloneq$ & $V^\star_P(\tE) - V^{\pi_P(\t)}_P(\tE)$\\
\midrule
\multicolumn{3}{l}{\emph{Information and identifiability}}\\
\midrule
$\cH_P(\t)$ & $\coloneq$ & $\nabla^2_\t V^\star_{P}(\t)$, information provided by $P$\\
$A^\t_P(\tau)$ & $\coloneq$ & feature advantage of $\pi_P(\t)$ along $\tau$ (Lemma~\ref{lemma:value_derivatives})\\
$\Sigma^\text{dyn}_P(\t)$ & $\coloneq$ & $\cov_P^{\pi_P(\t)}\br{\phi(\tau) - A^\t_P(\tau)}$, dynamics noise\\
$U_P$ & $\coloneq$ & $\im\cH_P(\t)$, identifiable subspace of $P$ (independent of $\t$)\\
$U$ & $\coloneq$ & $\sum_{P \in \cP} U_P$, joint identifiable subspace, of dimension $d_U$\\
$\Pi_P,\ \Pi_U$ & $\coloneq$ & orthogonal projectors onto $U_P$ and $U$\\
$G_P(\t_1, \t_2)$ & $\coloneq$ & $\int_0^1 (1-t)\, \cH_P\br{t\t_2 + (1-t)\t_1}\, \mathrm{d}t$\\
\midrule
\multicolumn{3}{l}{\emph{Estimation and confidence}}\\
\midrule
$\hat\phi_i$ & $\coloneq$ & $\frac1m\sum_{j=1}^m \phi(\tau_{i,j})$, empirical features at round $i$\\
$L_k(\t),\ \t_k$ & $\coloneq$ & empirical loss~\eqref{eq:minmaxirl}, and $\t_k\in\arg\min_{\t\in\T}L_k(\t)$\\
$\cH_k$ & $\coloneq$ & $\sum_{i=1}^k\cH_{P_i}(\tE)$, accumulated information\\
$\lambda_k$ & $\coloneq$ & $\lambda^U_{\min}(\cH_k)$, information in the least-informed direction\\
$\Sigma_P$ & $\coloneq$ & $\cov_P^{\piE_P}\br{\Pi_P\phi(\tau)}$\\
$d_\star$ & $\coloneq$ & $d_U \max_{P\in\cP}\norm{\cH_P(\tE)^{\dagger/2}\Sigma_P\cH_P(\tE)^{\dagger/2}}_2$, effective dimension\\
$R$ & $\coloneq$ & $\eta H (H+1) B_\phi$, pseudo self-concordance constant\\
$\beta_k^2(\delta)$ & $\coloneq$ & $\frac{16d_\star \log\br{3\eta d_\star km/\delta}}{(1-e^{-1})^2m}$,  squared confidence radius (Lemma~\ref{thm:confidencenew})\\
$n_\delta,\ n_0$ & $\coloneq$ & first round satisfying~\eqref{eq:n_min}, and warm-up length used in practice\\
\midrule
\multicolumn{3}{l}{\emph{Selection and transfer}}\\
\midrule
$k(P)$ & $\coloneq$ & $\sum_{i=1}^{k} \mathds{1}\bs{P_i = P}$, number of times $P$ was selected\\
$g_k(P)$ & $\coloneq$ & $\norm{\cH_k^{\dagger/2}\cH_P(\tE)\cH_k^{\dagger/2}}_2$, information gain of $P$\\
$\hat g_k(P)$ & $\coloneq$ & $\norm{\br{\sum_{i=1}^k\cH_i(\t_k)}^{\dagger/2}\cH_P(\t_k)\br{\sum_{i=1}^k\cH_i(\t_k)}^{\dagger/2}}_2$\\
$g_\perp(P)$ & $\coloneq$ & $\norm{\Pi_{U^\perp}\cH_P(\tE)\Pi_{U^\perp}}_2$\\
$\psi(x)$ & $\coloneq$ & $(xe^x - e^x + 1)/x^2$, with $\psi(0)\coloneq 1/2$\\
\midrule
\multicolumn{3}{l}{\emph{Linear algebra}}\\
\midrule
$\lambda^\cX_{\min}(X),\ \lambda^\cX_{\max}(X)$ & $\coloneq$ & $\min_{v\in \cX, \norm{v}_2=1} \norm{v}_X^2$, and $\max_{v\in \cX, \norm{v}_2=1} \norm{v}_X^2$\\
$A + B$ & $\coloneq$ & $\{a + b \mid a \in A,\, b \in B\}$\\
\bottomrule
\end{tabular}
\end{table}


\subsection{Preliminaries}

\paragraph{MDPs.}
Throughout this appendix, we work under the same regularized MDP setup introduced in Section~\ref{sec:formulation}. For a policy $\pi=(\pi_h)_{h=1}^H$, we define the state and state-action occupancies at each timestep $h\in\{1,\dots,H\}$ as
\[
    \nu^{\pi}_{h,P}(s) \coloneq \mathbb{P}_P^{\pi}\bs{s_h=s},
    \qquad
    \mu^{\pi}_{h,P}(s,a) \coloneq \nu^\pi_{h,P}(s)\,\pi_h(a \mid s).
\]
The regularized value then reads
\[
    V^\pi_P(r) = \sum_{h=1}^H \br{\ip{r}{\mu^\pi_{h,P}} + \frac{1}{\eta}\operatorname{Ent}_h(\pi)},
    \qquad
    \operatorname{Ent}_h(\pi) \coloneq -\sum_{s,a}\mu^\pi_{h,P}(s,a)\log\pi_h(a \mid s),
\]
where $\operatorname{Ent}_h(\pi)$ is the expected Shannon entropy of $\pi$ at timestep $h$ under $P$. The optimal policy in regularized MDPs has a softmax characterization~\citep{geist2019theory},
\begin{equation}
\label{eq:softmax_pi}
    \pi^\t_{h,P}(a \mid s) = \frac{\exp\br{\eta\, Q^\t_{h,P}(s,a)}}{\sum_{a' \in \cA}\exp\br{\eta\, Q^\t_{h,P}(s,a')}},
\end{equation}
where $Q^\t_{h,P}$ and $V^\t_{h,P}$ are the optimal regularized Q-function and value, defined by backward recursion from $V^\t_{H+1,P}\coloneq 0$,
\[
    Q^\t_{h,P}(s,a) \coloneq \ip{\t}{\phi(s,a)} + \sum_{s'\in\cS} T(s' \mid s,a)\, V^\t_{h+1,P}(s'),
    \qquad
    V^\t_{h,P}(s) \coloneq \frac{1}{\eta}\log\sum_{a\in\cA}\exp\br{\eta\, Q^\t_{h,P}(s,a)}.
\]

\paragraph{Eigenvalues.} The smallest eigenvalue of $X$ is denoted as $\lambda_{\min}(X) \coloneq \min_{\|v\|_2=1} \|v\|_X^2$ and the smallest eigenvalue restricted to a subspace $\cX\subset \R^d$ as $\lambda_{\min}^{\cX}(X) \coloneq \min_{\|v\|_2=1,\, v \in \cX} \|v\|_X^2$.

\paragraph{Features.} For convenience, we use the $\phi(\tau) = \sum_{h=1}^H \phi(s_h, a_h)$ to denote trajectory features, and $\phi_P(\t) = \mathbb{E}_P^{\pi_P(\t)}\bs{\phi(\tau)}$ to denote feature occupancies, with $\phi^\text{E}_P =\phi_P(\tE)$.

\paragraph {Fréchet derivatives.}
For a function $f : \mathbb{R}^d \to \mathbb{R}$ that is three times differentiable at $x \in \mathbb{R}^d$, we denote the first three directional derivatives along $u,v,w \in \mathbb{R}^d$ by
\begin{align*}
Df(x)[u]
:= \langle \nabla f(x), u \rangle,
\quad
D^2f(x)[u,v] \coloneq \ip{u}{\nabla^2 f(x) v},
\end{align*}
\[
D^3f(x)[u,v,w] \coloneq D\br{D^2f(x)[u, v]}[w].
\]
where $D^nf(x)$ denotes the $n$-th Fréchet derivative of $f$.

\section{Proofs of section~\ref{sec:conf}}
\label{apx:confidence_results}
\subsection{Reward estimation preliminaries}

In this paper, we consider minimizing the following aggregated empirical IRL loss to estimate rewards:
\[
L_k(\t) \coloneq \sum_{i=1}^k l_i(\t), \qquad l_i(\t) = V^\star_{P_i}(\Pi_{P_i}\t) - \ip{\Pi_{P_i}\t}{\hat\phi_i},
\]
where $\hat \phi_i = \frac{1}{m} \sum_{j=1}^m \phi(\tau_{i,j})$, and $\tau_{i,j}$ is the $j$-th trajectory generated by the expert at round $i$. We denote a minimizer over $\T$ by $\t_k \in \argmin_{\t\in\T} L_k(\t)$.
We furthermore define the population loss
\begin{equation}\label{eq:popp_loss}
L_k^\text{pop}(\t) = \sum_{i=1}^k \br{V^\star_{P_i}(\Pi_{P_i}\t) - \ip{\Pi_{P_i}\t}{\phi^\text{E}_{P_i}}}.
\end{equation}
Here $\phi^\text{E}_P =\mathbb{E}_P^{\pi^\text{E}_P}[\phi(\tau)]$.
Under adaptive dynamics selection, the population loss $L_k^\text{pop}$ is a sum of per-round conditional population losses. It is minimized at the expert parameters $\tE$.

Defining $S_k \coloneq \sum_{i=1}^k \Pi_{P_i}(\hat{\phi}_{i} - \phi^\text{E}_{P_i})$, the aggregated empirical loss decomposes as
\begin{equation*}
    L_k(\theta) = L^\text{pop}_k(\t) \;-\; \langle \theta,\, S_k\rangle.
\end{equation*}

Additionally, we introduce the following key quantities of our analysis:
\begin{align*}
\cH_P(\t) &:= \nabla^2_\t V^\star_P(\t),
& \cH_k
&:= \nabla^2 L_k(\theta^E) = \nabla^2 L^\text{pop}_k(\theta^E) = \sum_{i=1}^{k}\cH_{P_i}(\theta^E),
\\
\lambda_k &:= \lambda^{U}_{\min}(\cH_k),
&
R & \coloneq {\eta H(H+1)B_\phi}, \qquad R_k := \frac{R}{\sqrt{\lambda_k}},
\\
\rho_k &:= \|\theta_k - \theta^E\|_{\cH_k},
&
\omega_k &:= \|S_k\|_{\cH_k^\dagger}.
\end{align*}
Here, $\cH_k^\dagger$ denotes the Moore-Penrose inverse of $\cH_k$. The quantity $R_k$ is used only when $\lambda_k>0$. The stochastic self-concordance constant $R$ is justified in Lemma~\ref{lem:self_conc}.

Recall that $\Pi_P$ is the orthogonal projector onto $\im(\cH_P(0))$, which equals $\im(\cH_P(\t))$ for every $\t$ by Proposition~\ref{lemma:image}.
The parameter-independent kernel implies
\[
\Pi_P\,\cH_P(\Pi_P\t)\,\Pi_P = \cH_P(\t).
\]
Thus the projected loss has the same Hessian as the original loss, justifying the expression for $\cH_k$ above.

Moreover, projection ensures
\[
S_k \in \sum_{i=1}^k \im\bigl(\cH_{P_i}(\tE)\bigr) = \im(\cH_k).
\]
This support property allows the noise to be controlled in the inverse Hessian norm without requiring full identifiability.
\par

\subsection{Supporting lemmas}
\label{apx:reward_estimation_supporting_lemmas}

We present several technical lemmas laying out various properties of the optimal regularized value function and population loss~\ref{eq:popp_loss}. To make the dependence of the optimal value function on reward parameters explicit, we sometimes use the notation $V^\star_{\t, P} = V^\star_P(\t)$.

\begin{lemma}[Value derivatives~\citep{schlaginhaufen2026fastratesinversereinforcement}]\label{lemma:value_derivatives}
Let Assumption~\ref{as:linear_rewards} hold. Then, for any dynamics $P$ and any $\t,u,v \in \R^d$,
\begin{align*}
    DV^\star_P(\t)[v]
    &= \ip{v}{\phi_P(\t)},\\
    D^2V^\star_P(\t)[v,v]
    &= \norm{v}_{\cH_P(\t)}^2
     = \eta\,\mathbb{E}^{\pi_P(\t)}_P
       \bs{\ip{v}{A^\t_P(\tau)}^2},\\
    D^3V^\star_P(\t)[v,v,u]
    &= \eta^2\,\mathbb{E}^{\pi_P(\t)}_P
       \bs{\ip{v}{A^\t_P(\tau)}^2\ip{u}{A^\t_P(\tau)}},
\end{align*}
where $\phi_P(\t)\coloneq\mathbb{E}^{\pi_P(\t)}_P[\phi(\tau)]$ is the expected feature vector of the optimal policy, and $A^\t_P(\tau) \coloneq \sum_{h=1}^{H} A^\t_{h,P}(s_h,a_h)$ is the unregularized feature advantage, with
\[
A_{h,P}^\t(s,a) \coloneq \mathbb{E}_{P}^{\pi_P(\t)}\bs{\sum_{t=h}^{H} \phi(s_t,a_t) \;\bigg|\; s_h=s,\, a_h=a}
                - \mathbb{E}_{P}^{\pi_P(\t)}\bs{\sum_{t=h}^{H} \phi(s_t,a_t) \;\bigg|\; s_h=s}.
\]
In particular, $\mathbb{E}_P^{\pi_P(\t)}\bs{A^\t_P(\tau)} = 0$, since $\mathbb{E}_{a \sim \pi_{h,P}(\t)(\cdot \mid s)}\bs{A^\t_{h,P}(s,a)} = 0$ for every $h$ and $s$.
\end{lemma}
\begin{proof}
The three identities follow from \citet[Lemma~A.13]{schlaginhaufen2026fastratesinversereinforcement} under the linear reward structure of Assumption~\ref{as:linear_rewards}.
\end{proof}

\begin{corollary}[Deterministic transitions]
\label{cor:deterministic_transitions}
Let Assumption~\ref{as:linear_rewards} hold, and let $P$ have deterministic transitions and a deterministic initial state. Then, for any $\t,u,v \in \R^d$,
\begin{align*}
    DV^\star_{P}(\t)[v] &= \ip{v}{\phi_P(\t)},\\
    D^2V^\star_{P}(\t)[v, v] &= \norm{v}_{\cH_P(\t)}^2 = \eta\, \mathbb{E}^{\pi_P(\t)}_P \bs{\ip{v}{\phi(\tau) - \phi_P(\t)}^2},\\
    D^3V^\star_{P}(\t)[v, v, u] &= \eta^2\, \mathbb{E}^{\pi_P(\t)}_P \bs{\ip{v}{\phi(\tau) - \phi_P(\t)}^2 \ip{u}{\phi(\tau) - \phi_P(\t)}}.
\end{align*}
\end{corollary}

\begin{proof}
By Lemma~\ref{lemma:value_derivatives}, it suffices to show that $A^\t_P(\tau) = \phi(\tau) - \phi_P(\t)$. For $h\in\{1,\dots,H\}$, let
\[
\Phi_h(s) \coloneq \mathbb{E}_{P}^{\pi_P(\t)}\bs{\sum_{t=h}^{H} \phi(s_t,a_t) \;\bigg|\; s_h=s},
\qquad \Phi_{H+1}\coloneq 0.
\]
Since transitions are deterministic, $(s_h,a_h)$ determines $s_{h+1}$, so by the Markov property
\[
\mathbb{E}_{P}^{\pi_P(\t)}\bs{\sum_{t=h}^{H} \phi(s_t,a_t) \;\bigg|\; s_h, a_h} = \phi(s_h,a_h) + \Phi_{h+1}(s_{h+1}),
\]
and hence $A^\t_{h,P}(s_h,a_h) = \phi(s_h,a_h) + \Phi_{h+1}(s_{h+1}) - \Phi_h(s_h)$. Summing over $h$, the terms in $\Phi$ telescope:
\[
A^\t_P(\tau) = \sum_{h=1}^H \phi(s_h,a_h) - \Phi_1(s_1) = \phi(\tau) - \Phi_1(s_1).
\]
Since the initial state is deterministic, $\Phi_1(s_1) = \mathbb{E}_P^{\pi_P(\t)}[\phi(\tau)] = \phi_P(\t)$, which concludes the proof.
\end{proof}

\begin{lemma}[Covariance decomposition]
\label{lemma:feature_covariance}
Let Assumption~\ref{as:linear_rewards} hold. For any dynamics $P$ and $\t \in \R^d$,
\begin{equation} \label{eq:stochastic_covariance}
\cov_P^{\pi_P(\t)}\br{\phi(\tau)} = \eta^{-1}\cH_P(\t) + \Sigma_P^{\mathrm{dyn}}(\t),
\end{equation}
where, 
\[
\Sigma_P^{\mathrm{dyn}}(\t) \coloneq \cov_P^{\pi_P(\t)}\br{\phi(\tau)-A^\t_P(\tau)} \succeq 0.
\]
The matrix $\Sigma_P^{\mathrm{dyn}}(\t)$ captures the randomness of the transitions and of the initial state.
\end{lemma}

\begin{proof}
Since $\mathbb{E}_P^{\pi_P(\t)}\bs{A^\t_P(\tau)} = 0$, Lemma~\ref{lemma:value_derivatives} gives $\cov_P^{\pi_P(\t)}\br{A^\t_P(\tau)} = \mathbb{E}_P^{\pi_P(\t)}\bs{A^\t_P(\tau)A^\t_P(\tau)^\top} = \eta^{-1}\cH_P(\t)$. Moreover, in the decomposition
\[
\phi(\tau) - \phi_P(\t) = A^\t_P(\tau) + \br{\phi(\tau) - A^\t_P(\tau) - \phi_P(\t)},
\]
the two terms are uncorrelated by \citet[Lemma~A.7]{schlaginhaufen2026fastratesinversereinforcement}. Taking covariances on both sides proves~\eqref{eq:stochastic_covariance}.
\end{proof}

\begin{proposition}[Hessian upper bound]
\label{proposition:hessian_ub}
Let Assumption~\ref{as:linear_rewards} hold. For any dynamics $P$ and $\t \in \R^d$, $\norm{\cH_P(\t)}_2 \le \eta H^2 B_{\phi}^2$.
\end{proposition}

\begin{proof}
Let $v\in\R^d$ with $\norm{v}_2=1$. By Lemma~\ref{lemma:value_derivatives} and the covariance decomposition of Lemma~\ref{lemma:feature_covariance},
\begin{align*}
\norm{v}_{\cH_P(\t)}^2 &= \eta\,\mathbb{E}_P^{\pi_P(\t)} \bs{\ip{v}{A^\t_P(\tau)}^2}
\le \eta\,\var_P^{\pi_P(\t)} \br{\ip{v}{\phi(\tau)}}
\le \eta\,\mathbb{E}_P^{\pi_P(\t)} \bs{\ip{v}{\phi(\tau)}^2}
\le \eta H^2 B_\phi^2,
\end{align*}
where the last step uses $\norm{\phi(\tau)}_2\le \sum_{h=1}^H \norm{\phi(s_h,a_h)}_2 \le H B_\phi$. Since $\cH_P(\t)$ is positive semidefinite, $\norm{\cH_P(\t)}_2 = \max_{\norm{v}_2=1}\norm{v}_{\cH_P(\t)}^2$, which proves the claim.
\end{proof}


\begin{proposition}[Hessian image~\citep{schlaginhaufen2026fastratesinversereinforcement}]
\label{lemma:image}
Let Assumption~\ref{as:linear_rewards} hold and let $P$ be any dynamics. Then $\ker\br{\cH_P(\t)}$ consists of the directions $v\in\R^d$ whose reward perturbation $r_v(s,a)\coloneq\ip{v}{\phi(s,a)}$ preserves the optimal trajectory law under $P$, and is independent of $\t$. In particular, $U_P = \im\br{\cH_P(\t)} = \im\br{\cH_P(0)}$ for all $\t\in\R^d$.
\end{proposition}
\begin{proof}
By Lemma~\ref{lemma:value_derivatives}, for any $v\in\R^d$,
\begin{align*}
v\in\ker\br{\cH_P(\t)}
&\iff \mathbb{E}_P^{\pi_P(\t)} \bs{\ip{v}{A^\t_P(\tau)}^2}=0\\
&\iff \ip{v}{A^\t_P(\tau)}=0 \quad \mathbb{P}_P^{\pi_P(\t)}\text{-almost surely}.
\end{align*}
By linearity, $\ip{v}{A^\t_P(\tau)}$ is the cumulative unregularized advantage of $r_v$ under $\pi_P(\t)$. By \citet[Corollary~A.16]{schlaginhaufen2026fastratesinversereinforcement}, it vanishes almost surely if and only if $r_v$ preserves the optimal trajectory law under $P$, a condition that does not depend on $\t$ since soft-optimal policies have full support. Hence $\ker\br{\cH_P(\t)} = \ker\br{\cH_P(0)}$ for all $\t$. Since $\cH_P(\t)$ is symmetric, $\im\br{\cH_P(\t)} = \ker\br{\cH_P(\t)}^\perp$, which concludes the proof.
\end{proof}

\begin{lemma}[Pseudo self-concordance]
\label{lem:self_conc}
Let Assumption~\ref{as:linear_rewards} hold, and define $R \coloneq \eta H (H+1)B_\phi$. For any dynamics $P$ and $\t, u, v \in \R^d$, the optimal regularized value function satisfies
\begin{equation*}
    \bigl|D^3 V^\star_{P}(\t)[v, v, u]\bigr|
    \leq R \,\norm{u}_{\Pi_P}\, D^2 V^\star_{P}(\t) [v, v].
\end{equation*}
\end{lemma}
\begin{proof}
By Lemma~\ref{lemma:value_derivatives} and Proposition~\ref{lemma:image}, $\ip{w}{A^\t_P(\tau)}=0$ almost surely for each $w$ in a basis of $\ker\cH_P(\t) = U_P^\perp$. Hence $A^\t_P(\tau)\in U_P$ almost surely, and
\[
    \ip{u}{A^\t_P(\tau)} = \ip{\Pi_Pu}{A^\t_P(\tau)} \qquad\text{almost surely}.
\]
At each step $h$, the feature advantage $A^\t_{h,P}$ is the difference of two conditional expectations of the remaining feature sum, each of norm at most $(H-h+1)B_\phi$. Therefore,
\[
    \norm{A^\t_P(\tau)}_2 \le 2B_\phi\sum_{h=1}^H(H-h+1) = H(H+1)B_\phi
    \quad\text{almost surely}.
\]
Using Lemma~\ref{lemma:value_derivatives} and the Cauchy--Schwarz inequality,
\begin{align*}
    \bigl|D^3V^\star_P(\t)[v,v,u]\bigr| &= \eta^2 \left| \mathbb{E}_P^{\pi_P(\t)} \bs{\ip{v}{A^\t_P(\tau)}^2 \ip{\Pi_Pu}{A^\t_P(\tau)}}\right|\\
    &\le \eta^2 H(H+1)B_\phi\,\norm{\Pi_Pu}_2\,\mathbb{E}_P^{\pi_P(\t)}\bs{\ip{v}{A^\t_P(\tau)}^2}\\
    &= R\,\norm{u}_{\Pi_P}\,D^2V^\star_P(\t)[v,v].
\end{align*}
\end{proof}

\begin{lemma}[Curvature stability]
\label{lem:curvature_stability}
Let Assumption~\ref{as:linear_rewards} hold and let $P$ be any dynamics. Fix $\t_1, \t_2 \in \R^d$ and set $a \coloneq R\norm{\t_2 - \t_1}_{\Pi_P}$. Define
\[
    \psi(x) \coloneq \int_0^1 t\, e^{xt}\,\mathrm{d}t =
    \begin{cases}
        \dfrac{xe^x - e^x + 1}{x^2}, & x \neq 0,\\[1ex]
        1/2, & x = 0,
    \end{cases}
    \qquad
    G_P(\t_1, \t_2) \coloneq \int_0^1 (1-t)\,\cH_P\br{t\t_2 + (1-t)\t_1}\,\mathrm{d}t.
\]
Then:
\begin{enumerate}
    \item $e^{-a}\,\cH_P(\t_2) \preceq \cH_P(\t_1) \preceq e^{a}\,\cH_P(\t_2)$;
    \item $\psi(-a)\,\cH_P(\t_2) \preceq G_P(\t_1, \t_2) \preceq \psi(a)\,\cH_P(\t_2)$;
    \item $e^{-a}\psi(a)\,\cH_P(\t_1) \preceq G_P(\t_1, \t_2) \preceq e^{a}\psi(-a)\,\cH_P(\t_1)$.
\end{enumerate}
\end{lemma}

\begin{proof}
Let $\bar\t(t) \coloneq t\t_2 + (1-t)\t_1$ for $t \in [0,1]$, and recall that $D^2V^\star_P(\t)[v,v] = v^\top \cH_P(\t)\, v$.

\textit{Reduction to $v \in U_P \setminus \{0\}$.}
Let $v \in \R^d$. By Proposition~\ref{lemma:image}, $(I - \Pi_P)v \in \ker\br{\cH_P(\t)}$ for every $\t \in \R^d$, so by symmetry of $\cH_P(\t)$,
\[
    v^\top \cH_P(\t)\, v = (\Pi_P v)^\top \cH_P(\t)\, (\Pi_P v)
    \qquad \text{for all } \t \in \R^d,
\]
and integrating this identity gives the same for $G_P(\t_1,\t_2)$. If $\Pi_P v = 0$, every quadratic form in the claim vanishes. It therefore suffices to prove the scalar inequalities for $v \in U_P \setminus \{0\}$, which we fix from now on.

\textit{Pointwise bounds.}
By Proposition~\ref{lemma:image}, $q(t) \coloneq D^2V^\star_P(\bar\t(t))[v,v] > 0$ for all $t \in [0,1]$. Since $q'(t) = D^3V^\star_P(\bar\t(t))[v,v,\t_2-\t_1]$, Lemma~\ref{lem:self_conc} gives
\[
    \abs{\frac{\mathrm{d}}{\mathrm{d}t} \log q(t)} \le a
    \qquad \text{for all } t \in [0,1].
\]
Hence $\abs{\log q(t) - \log q(s)} \le \abs{t-s}\,a$ for all $s, t \in [0,1]$. Taking $s = 1$ and $s = 0$ and exponentiating, we get, for all $t \in [0,1]$,
\begin{align}
    e^{-(1-t)a}\, q(1) &\le q(t) \le e^{(1-t)a}\, q(1), \label{eq:pointwise_t2}\\
    e^{-ta}\, q(0) &\le q(t) \le e^{ta}\, q(0). \label{eq:pointwise_t1}
\end{align}

\textit{Item 1.}
This is~\eqref{eq:pointwise_t2} at $t = 0$.

\textit{Items 2 and 3.}
Note that $v^\top G_P(\t_1,\t_2)\, v = \int_0^1 (1-t)\, q(t)\,\mathrm{d}t$. Multiplying~\eqref{eq:pointwise_t2} and~\eqref{eq:pointwise_t1} by $(1-t) \ge 0$ and integrating over $[0,1]$ preserves the inequalities. With the substitution $s = 1-t$ and the definition of $\psi$, the resulting weights are
\[
    \int_0^1 (1-t)\, e^{\pm(1-t)a}\,\mathrm{d}t = \int_0^1 s\, e^{\pm as}\,\mathrm{d}s = \psi(\pm a),
    \qquad
    \int_0^1 (1-t)\, e^{\pm ta}\,\mathrm{d}t = e^{\pm a}\int_0^1 s\, e^{\mp as}\,\mathrm{d}s = e^{\pm a}\psi(\mp a).
\]
This yields items 2 and 3, respectively.
\end{proof}

\begin{lemma}
\label{lemma:loss_self_conf}
Let Assumption~\ref{as:linear_rewards} hold, and define $\chi(x) \coloneq \frac{e^x-1}{x}$ for $x\neq0$, with $\chi(0)\coloneq1$. Then, for any $k\ge1$ with $\lambda_k>0$,
\[
    \chi\br{-R_k \rho_k}\, \rho_k^2 \le \ip{\t_k -\tE}{\nabla L^\text{pop}_k(\t_k) - \nabla L^\text{pop}_k(\tE)} \le \chi\br{R_k \rho_k}\, \rho_k^2.
\]
\end{lemma}
\begin{proof}
Let $u \coloneq \t_k - \tE$ and $\bar\t(t)\coloneq \tE + t\,u$ for $t\in[0,1]$. Since $\cH_k$ vanishes on $U^\perp$ and $\norm{w}^2_{\cH_k}\ge\lambda_k\norm{w}_2^2$ for $w\in U$, we have $\cH_k\succeq\lambda_k\Pi_U$. As $U_{P_i}\subseteq U$, this gives for every $i\le k$
\[
    a_i \coloneq R\norm{u}_{\Pi_{P_i}} \le R\norm{\Pi_U u}_2 \le R\frac{\norm{u}_{\cH_k}}{\sqrt{\lambda_k}} = R_k \rho_k.
\]
Applying~\eqref{eq:pointwise_t1} from the proof of Lemma~\ref{lem:curvature_stability} to each $P_i$ with $\t_1 = \tE$ and $\t_2 = \t_k$, and summing over $i$, we get for any $v \in \R^d$ and $t\in[0,1]$
\begin{equation}
\label{eq:exp_conc_loss}
    e^{-t R_k\rho_k} \norm{v}^2_{\cH_k} \le D^2 L^\text{pop}_k(\bar\t(t))[v,v] \le e^{t R_k\rho_k} \norm{v}^2_{\cH_k},
\end{equation}
where we used $a_i\le R_k\rho_k$ and $D^2 L^\text{pop}_k(\t) = \sum_{i=1}^k\cH_{P_i}(\t)$.

By the fundamental theorem of calculus,
\[
    \ip{u}{\nabla L^\text{pop}_k(\t_k) - \nabla L^\text{pop}_k(\tE)} = \int_0^1 D^2 L^\text{pop}_k(\bar\t(t))[u,u]\,\mathrm{d}t.
\]
Sandwiching the integrand using~\eqref{eq:exp_conc_loss} with $v = u$, and using $\int_0^1 e^{\pm xt}\,\mathrm{d}t = \chi(\pm x)$ and $\norm{u}_{\cH_k}=\rho_k$, gives the result.
\end{proof}

\subsection{Ellipsoidal confidence sets}

\begin{remark}\label{rem:deviation_span_remark}
For any $u \in \R^d$,
\begin{equation}\label{eq:deviation_span_remark}
\ip{S_k}{u} = \ip{(\cH_k^\dagger)^{1/2} S_k}{\cH_k^{1/2}u}.
\end{equation}
Indeed, let $v \in \ker(\cH_k)$. Since $\cH_k = \sum_{i=1}^k \cH_{P_i}(\tE)$ is a sum of positive semidefinite matrices, $v^\top\cH_{P_i}(\tE)\,v = 0$ and hence $v \in \ker\br{\cH_{P_i}(\tE)}$ for every $i$. By Proposition~\ref{lemma:image},
\[
\ker\br{\cH_{P_i}(\tE)} = U_{P_i}^{\perp} = \ker(\Pi_{P_i}),
\]
so $\Pi_{P_i}v=0$ for every $i$. By definition of $S_k$ and symmetry of the orthogonal projectors,
\[
\ip{v}{S_k} = \sum_{i=1}^k\ip{v}{\Pi_{P_i} \br{\hat\phi_i-\phi^\mathrm{E}_{P_i}}} = \sum_{i=1}^k \ip{\Pi_{P_i}v}{\hat\phi_i-\phi^\mathrm{E}_{P_i}} = 0.
\]
Therefore $S_k \in \ker(\cH_k)^\perp = \im(\cH_k)$. Since $\cH_k^{\dagger/2}\cH_k^{1/2}$ is the orthogonal projector onto $\im(\cH_k)$, the right-hand side of~\eqref{eq:deviation_span_remark} equals $\ip{S_k}{u}$.
\end{remark}

To proceed, we use the first-order optimality of $\t_k$ for $L_k$ over $\T$ and of $\tE$ for $L_k^{\mathrm{pop}}$ to derive a deterministic localization bound.
\begin{lemma}[Deterministic localization bound]
\label{lem:localization}
Let Assumption~\ref{as:linear_rewards} hold and $\tE\in\T$. For all $k\ge1$ with $\lambda_k > 0$,
\begin{equation}\label{eq:chi_ineq}
    \chi(-R_k\,\rho_k)\,\rho_k \;\leq\; \omega_k.
\end{equation}
Additionally, on the localization event
\begin{equation}\label{eq:local_event}
    \cE_k \coloneq \Bigl\{\omega_k \leq \frac{1 - e^{-1}}{R_k}\Bigr\},
\end{equation}
we have $R_k \rho_k \leq 1$ and
\begin{equation}\label{eq:rho_eta}
    \rho_k \leq \frac{1}{1-e^{-1}}\,\omega_k.
\end{equation}
\end{lemma}

\begin{proof}
Since $\t_k$ minimizes $L_k$ over $\T$, the first-order optimality condition gives
\[
    \ip{\nabla L_k(\t_k)}{\t - \t_k} \geq 0
    \quad \forall\, \t \in \T.
\]
By well-specification and feature matching for the projected population loss, $\nabla L_k^{\mathrm{pop}}(\tE)=0$. Using the decomposition $\nabla L_k(\t_k) = \nabla L^\text{pop}_k(\t_k) - S_k$ and taking $\t=\tE$, we have
\[
    \ip{\nabla L^\text{pop}_k(\t_k) - S_k}{\tE - \t_k} \geq 0.
\]
Rearranging and subtracting $\ip{\nabla L^\text{pop}_k(\tE)}{\t_k - \tE} = 0$,
\begin{equation}\label{eq:rearrange}
    \ip{\nabla L^\text{pop}_k(\t_k) - \nabla L^\text{pop}_k(\tE)}{\t_k - \tE}
    \leq \ip{S_k}{\t_k - \tE}.
\end{equation}
By Remark~\ref{rem:deviation_span_remark} and the Cauchy--Schwarz inequality, the right-hand side of~\eqref{eq:rearrange} satisfies
\begin{equation}\label{eq:cauchy}
    \ip{S_k}{\t_k - \tE}
    = \ip{(\cH_k^\dagger)^{1/2} S_k}{\cH_k^{1/2}(\t_k - \tE)}\leq \omega_k\,\rho_k.
\end{equation}
Combining the lower bound of Lemma~\ref{lemma:loss_self_conf} with~\eqref{eq:rearrange} and~\eqref{eq:cauchy},
\[
    \chi(-R_k\rho_k)\,\rho_k^2 \leq \omega_k\,\rho_k.
\]
Dividing by $\rho_k$ (if $\rho_k > 0$; the bound is trivial otherwise) yields~\eqref{eq:chi_ineq}.

Finally, on the event $\cE_k$, inequality~\eqref{eq:chi_ineq} gives
\[
    \frac{1-e^{-R_k\rho_k}}{R_k} = \chi(-R_k\,\rho_k)\,\rho_k \leq \omega_k \leq \frac{1 - e^{-1}}{R_k},
\]
hence $1-e^{-R_k\rho_k} \le 1-e^{-1}$, i.e.\ $R_k\rho_k \le 1$. Since $\chi$ is increasing, $\chi(-R_k\rho_k) \ge \chi(-1) = 1-e^{-1}$, and therefore
\[
    (1-e^{-1})\,\rho_k \leq \chi(-R_k\rho_k)\,\rho_k \leq \omega_k,
\]
which gives~\eqref{eq:rho_eta}.
\end{proof}

In the next step, we are bounding the excess risk on the event $\cE$.

\begin{lemma}[Excess risk bound on $\cE_k$]
\label{lem:excess_risk}
Let Assumption~\ref{as:linear_rewards} hold and $\tE\in\T$. Fix $k\ge1$ with $\lambda_k>0$. On the event $\cE_k$ defined in~\eqref{eq:local_event},
\begin{equation}\label{eq:excess_risk}
    L^\text{pop}_k(\t_k) - L^\text{pop}_k(\tE) \leq \frac{1}{1-e^{-1}}\,\omega_k^2.
\end{equation}
\end{lemma}

\begin{proof}
Since $\t_k$ minimizes $L_k$ over $\T$ and $\tE \in \T$, we have $L_k(\t_k) \leq L_k(\tE)$. Using $L_k(\t) = L^\text{pop}_k(\t) - \ip{\t}{S_k}$,
\begin{align*}
    L^\text{pop}_k(\t_k) - L^\text{pop}_k(\tE)
    &= \ip{\t_k - \tE}{S_k}
     + \br{L_k(\t_k) - L_k(\tE)}\\
    &\leq \ip{\t_k - \tE}{S_k}.
\end{align*}
By~\eqref{eq:cauchy} and, on $\cE_k$, Lemma~\ref{lem:localization},
\[
    \ip{\t_k - \tE}{S_k}
    \leq \omega_k\,\rho_k
    \leq \frac{1}{1-e^{-1}}\,\omega_k^2.
\]
\end{proof}

Fix the number of trajectories per round $m\in\mathbb N^+$ and $\delta\in(0,1)$. Recall that $U = \sum_{P\in\cP}U_P$ has dimension $d_U\le d$. We assume $d_U\ge1$: if $U=\{0\}$, all projected scores and Hessians vanish and there is nothing to estimate. At each round $k\ge1$, $P_k$ is selected based on the history, and $m$ fresh, conditionally independent expert trajectories are drawn under $P_k$. Recall
\begin{align*}
 \cH_k &\coloneq \sum_{i=1}^k\cH_{P_i}(\tE), &\lambda_k&\coloneq \min_{v\in U,\ \norm{v}_2=1}\norm{v}^2_{\cH_k},\\
 \Sigma_P &\coloneq \cov_{P}^{\piE_P}\br{\Pi_P\phi(\tau)}, &d_\star&\coloneq d_U\,\max_{P\in\cP}\norm{\cH_P(\tE)^{\dagger/2}\,\Sigma_P\,\cH_P(\tE)^{\dagger/2}}_2.
\end{align*}
Note that $\lambda_k = 0$ until the selected dynamics cover all of $U$.

Since $\Sigma_P$ is supported on $U_P = \im\cH_P(\tE)$, the definition of $d_\star$ gives
\begin{equation}
 \label{eq:covariance_curvature_constant}
 \Sigma_P \;\preceq\; \frac{d_\star}{d_U}\,\cH_P(\tE) \qquad \text{for every } P\in\cP.
\end{equation}
Projecting the covariance decomposition~\eqref{eq:stochastic_covariance} onto $U_P$ gives
\[
 \Sigma_P=\eta^{-1}\cH_P(\tE) +\Pi_P\Sigma_P^{\mathrm{dyn}}(\tE)\Pi_P \succeq\eta^{-1}\cH_P(\tE).
\]
Thus $d_\star\ge d_U/\eta$, with equality for deterministic transitions and initial state (Corollary~\ref{cor:deterministic_transitions}). The ratio $d_\star/d_U$ measures the largest sampling variance relative to curvature along an identifiable direction, and stochastic transitions can only increase it.

We define the analysis regularization and logarithmic factor
\begin{equation}
 \label{eq:self_normalized_log_factor}
 \gamma \coloneq \frac{4\,d_U H^2B_\phi^2}{d_\star\, m},
 \qquad
 \ell_{k,\delta} \coloneq \log\frac1\delta + d_U\log2 + \frac12\log\det\br{I+\frac{\cH_k}{\gamma}} + \frac14.
\end{equation}
The matrix inside the determinant is the identity on $U^\perp$, so only directions in $U$ contribute.
For the subsequent selection guarantees, we fix the budget $n$.

\begin{lemma}[Anytime self-normalized concentration]
\label{lem:omega_concentration}
Let Assumption~\ref{as:linear_rewards} hold, and define
\[
 X_{i,j}\coloneq \frac1m\Pi_{P_i}\br{\phi(\tau_{i,j})-\phi_{P_i}^{\mathrm E}}, \quad S_k\coloneq \sum_{i=1}^k\sum_{j=1}^mX_{i,j}, \quad \omega_k\coloneq \norm{S_k}_{\cH_k^\dagger}.
\]
With probability at least $1-\delta$, simultaneously for all $k\ge1$,
\begin{equation}
 \label{eq:mixture_regularized_bound}
 \norm{S_k}_{(\cH_k+\gamma I)^{-1}} \le 2\sqrt{\frac{d_\star\,\ell_{k,\delta}}{d_U\,m}} +\frac{8HB_\phi\ell_{k,\delta}}{m\sqrt{\lambda_k+\gamma}}.
\end{equation}
On the same event, every round $k$ with $\lambda_k\ge \gamma$ satisfies
\begin{equation}
 \label{eq:omega_bound}
 \omega_k \le 2\sqrt{\frac{2d_\star\,\ell_{k,\delta}}{d_U\,m}} +\frac{8HB_\phi\ell_{k,\delta}}{m\sqrt{\lambda_k}}.
\end{equation}
\end{lemma}

\begin{proof}
Since the Hessian images do not depend on the parameters, $U=\sum_{P\in\cP}\im\cH_P(0)$ is fixed before sampling. In this proof, we work in a fixed orthonormal basis of $U$, with all vectors and matrices restricted to $U$ (and $I$ the identity on $U$) unless otherwise indicated. Each increment $X_{i,j}$ lies in $U$, as it is projected onto $U_{P_i}\subseteq U$. We use the shorthands
\[
 c\coloneq\frac{d_\star}{d_U\,m}, \qquad r\coloneq\frac{m}{2HB_\phi}, \qquad\text{so that } \gamma c = r^{-2} \text{ by~\eqref{eq:self_normalized_log_factor}}.
\]

\emph{Step 1: supermartingale.} Let $\cG_{i,j-1}$ contain all information available before trajectory $j$ of round $i$, including $P_i$ and any learner randomization used to select it. Fresh sampling gives
\[
 \mathbb E[X_{i,j}\mid\cG_{i,j-1}]=0, \quad \norm{X_{i,j}}_2\le\frac{2HB_\phi}{m} = \frac1r,
\]
and, by~\eqref{eq:covariance_curvature_constant},
\[
 \mathbb E[X_{i,j}X_{i,j}^{\top}\mid\cG_{i,j-1}] =\frac{\Sigma_{P_i}}{m^2} \preceq \frac{c}{m}\,\cH_{P_i}(\tE).
\]
For every fixed $z\in U$ with $\norm{z}_2\le r$, we have $|z^{\top}X_{i,j}|\le1$. Using $e^x\le1+x+x^2$ for $|x|\le1$, we obtain
\[
 \mathbb E[e^{z^{\top}X_{i,j}}\mid\cG_{i,j-1}]
 \le1+\mathbb E[(z^{\top}X_{i,j})^2\mid\cG_{i,j-1}]
 \le\exp\!\left(\frac{c}{m}\, z^{\top}\cH_{P_i}(\tE)z\right).
\]
Iterating this bound over the $m$ trajectories of each round shows that
\[
 M_k(z)\coloneq \exp\!\br{ z^{\top}S_k - c\, z^{\top}\cH_kz }
\]
is a nonnegative supermartingale with $M_0(z)=1$.

\emph{Step 2: mixture and Ville's inequality.} Let $\nu_0$ be the uniform distribution on $\{z\in U:\norm{z}_2\le r\}$. The mixture $\int M_k(z)\,\nu_0(\mathrm dz)$ is a nonnegative supermartingale starting at one. By Ville's inequality, with probability at least $1-\delta$,
\[
 \int M_k(z)\,\nu_0(\mathrm dz)\le1/\delta \quad\text{for all }k\ge0.
\]
On this event, the Donsker--Varadhan variational inequality gives, for every round $k$ and every probability distribution $\nu\ll\nu_0$,
\begin{equation}
 \label{eq:mixture_variational}
 \mathbb E_{z\sim\nu}\!\bs{z^{\top}S_k - c\, z^{\top}\cH_kz } \le \operatorname{KL}(\nu\|\nu_0)+\log\frac1\delta.
\end{equation}
Since this holds for all such $\nu$ simultaneously, $\nu$ may depend on the data.

\emph{Step 3: choice of $\nu$.} Fix $k$ and define the positive definite matrix on $U$
\[
 W_k\coloneq c\,\cH_k + r^{-2} I \;=\; c\br{\cH_k+\gamma I}.
\]
In particular, $\lambda_{\min}(W_k) = c\,(\lambda_k+\gamma) \ge r^{-2}$. If $S_k=0$, the claimed bounds at this round hold trivially. Otherwise, set
\[
 \alpha_k\coloneq \min\!\left\{\frac12,\,\frac{r\sqrt{\lambda_{\min}(W_k)}}{2\norm{S_k}_{W_k^{-1}}}\right\},
\]
and let $\nu_k$ be uniform on the ellipsoid
\[
 \left\{\alpha_kW_k^{-1}S_k+\frac12W_k^{-1/2}u:u\in U,\ \norm{u}_2\le1\right\}.
\]
Its center and radius satisfy
\[
 \norm{\alpha_kW_k^{-1}S_k}_2\le\frac{\alpha_k\norm{S_k}_{W_k^{-1}}}{\sqrt{\lambda_{\min}(W_k)}}\le\frac{r}{2}, \quad \frac12\norm{W_k^{-1/2}}_2\le\frac{r}{2},
\]
where the second bound uses $\lambda_{\min}(W_k)\ge r^{-2}$. Hence $\nu_k\ll\nu_0$. Its mean and covariance are
\[
 \mathbb E_{\nu_k}[z]=\alpha_kW_k^{-1}S_k, \quad \cov_{\nu_k}(z) =\frac{W_k^{-1}}{4(d_U+2)},
\]
and, since $r^2W_k = I+\cH_k/\gamma$, the ratio of support volumes gives
\[
 \operatorname{KL}(\nu_k\|\nu_0) = d_U\log2 +\frac12\log\det\nolimits_U\!\br{I+\frac{\cH_k}{\gamma}}.
\]
This restricted determinant equals the full-space determinant in~\eqref{eq:self_normalized_log_factor}, since $\cH_k$ vanishes on $U^\perp$.

\emph{Step 4: solving for $S_k$.} Since $0\preceq c\,\cH_k\preceq W_k$,
\begin{align*}
 \mathbb E_{\nu_k}\!\bs{ z^{\top}S_k - c\, z^{\top}\cH_kz }
 &=\alpha_k\norm{S_k}_{W_k^{-1}}^2- c\,\alpha_k^2\, S_k^{\top}W_k^{-1}\cH_kW_k^{-1}S_k -\frac{c\,\operatorname{tr}(\cH_kW_k^{-1})} {4(d_U+2)}\\
 &\ge \alpha_k(1-\alpha_k)\norm{S_k}_{W_k^{-1}}^2-\frac14.
\end{align*}
Substituting into~\eqref{eq:mixture_variational} yields
$\alpha_k(1-\alpha_k)\norm{S_k}_{W_k^{-1}}^2 \le\ell_{k,\delta}$.
If $\alpha_k=1/2$, then $\norm{S_k}_{W_k^{-1}}\le2\sqrt{\ell_{k,\delta}}$. Otherwise, $1-\alpha_k\ge1/2$ and the definition of $\alpha_k$ gives
$\norm{S_k}_{W_k^{-1}} \le 4\ell_{k,\delta}/(r\sqrt{\lambda_{\min}(W_k)})$.
Combining the two cases,
\[
 \norm{S_k}_{W_k^{-1}} \le 2\sqrt{\ell_{k,\delta}} +\frac{4\ell_{k,\delta}} {r\sqrt{\lambda_{\min}(W_k)}}.
\]
Since $\norm{S_k}_{W_k^{-1}} = c^{-1/2}\norm{S_k}_{(\cH_k+\gamma I)^{-1}}$ and $\lambda_{\min}(W_k) = c\,(\lambda_k+\gamma)$, rescaling proves~\eqref{eq:mixture_regularized_bound}. The norm agrees with its full-space version because $S_k\in U$.

\emph{Step 5: removing the regularization.} If $\lambda_k\ge \gamma$, then $\cH_k\succeq\lambda_k I$ on $U$, so $\cH_k^{-1}\preceq(1+\gamma/\lambda_k)(\cH_k+\gamma I)^{-1}$ there, and
\[
 \omega_k \le \sqrt{1+\frac{\gamma}{\lambda_k}}\, \norm{S_k}_{(\cH_k+\gamma I)^{-1}}.
\]
Applying~\eqref{eq:mixture_regularized_bound} and $1+\gamma/\lambda_k\le2$ gives~\eqref{eq:omega_bound}. The linear term simplifies exactly, because
$\sqrt{1+\gamma/\lambda_k}\,/\sqrt{\lambda_k+\gamma} =1/\sqrt{\lambda_k}$.
All bounds hold on the same event for every round $k$.
\end{proof}

Let $\Theta$ be convex with $\tE\in\Theta$, and let $\theta_k$
minimize the {projected} empirical loss over $\Theta$.
Let $R\ge2\eta HB_\phi$ be a valid pseudo-self-concordance constant
from Lemma~\ref{lem:self_conc}.
{For the stochastic setting, the established choice is
$R=\eta H(H+1)B_\phi$.}
The deterministic localization lemma remains a pathwise result and
requires no additional concentration argument.

\confidencenew*
\begin{proof}
Work on the event from Lemma~\ref{lem:omega_concentration}. Throughout, we use that $d_\star \ge d_U/\eta$ and $R = \eta H(H+1)B_\phi \ge 2\eta HB_\phi$, so that
\begin{equation}
\label{eq:R_dstar}
    R\,\frac{d_\star}{d_U} \;\ge\; \frac{R}{\eta} \;\ge\; 2HB_\phi.
\end{equation}
The argument relies on the condition
\begin{equation}
\label{eq:lambdan_warmup}
    \lambda_k \;\ge\; \frac{32 R^2 d_\star\,\ell_{k,\delta}}{(1-e^{-1})^2\,d_U\, m}.
\end{equation}

\emph{Step 1: the warm-up condition implies~\eqref{eq:lambdan_warmup}.} By Proposition~\ref{proposition:hessian_ub}, every eigenvalue of $\cH_k$ is at most $k\eta H^2B_\phi^2$. With $\gamma = 4d_UH^2B_\phi^2/(d_\star m)$, every eigenvalue of $\cH_k/\gamma$ is thus at most $x/4$, where $x \coloneq km\,\eta d_\star/d_U \ge 1$. Since the determinant in $\ell_{k,\delta}$ only involves the $d_U$ dimensions of $U$,
\[
    \ell_{k,\delta}
    \;\le\; \log\tfrac1\delta + d_U\log 2 + \tfrac{d_U}{2}\log\br{1+\tfrac{x}{4}} + \tfrac14
    \;=\; \log\tfrac1\delta + \tfrac{d_U}{2}\log(4+x) + \tfrac14.
\]
Since $x\ge1$, we have $9x^2/(4+x)\ge 9/5$ and $\tfrac12\log\tfrac95\ge\tfrac14$, hence $\tfrac{d_U}{2}\log(4+x)+\tfrac14 \le d_U\log(3x)$. With $\log\tfrac1\delta\le d_U\log\tfrac1\delta$ and $x\le km\,\eta d_\star$, this gives
\begin{equation}
\label{eq:ell_bound}
    \ell_{k,\delta} \;\le\; d_U\log(3x/\delta) \;\le\; d_U\log(3\eta d_\star km/\delta).
\end{equation}
In particular, $\ell_{k,\delta} \le d_U\log(3\eta d_\star nm/\delta)$ for every $1\le k\le n$. Since $\cH_k\succeq\cH_{n_\delta}$ for $k\ge n_\delta$, the warm-up condition~\eqref{eq:n_min} implies
\[
    \lambda_k \ge \lambda_{n_\delta} \ge \frac{32R^2d_\star}{(1-e^{-1})^2m} \log\frac{3\eta d_\star nm}{\delta} \ge \frac{32R^2d_\star\ell_{k,\delta}}{(1-e^{-1})^2d_Um} \quad \text{for every }n_\delta\le k\le n.
\]
Thus~\eqref{eq:lambdan_warmup} holds throughout the remaining budget, independently of the dynamics selected after the warm-up.

\emph{Step 2: localization.} Fix a round $n_\delta\le k\le n$. Using $\ell_{k,\delta}\ge1/4$, the right-hand side of~\eqref{eq:lambdan_warmup} is at least $8R^2 d_\star/(d_U m)$, which is at least $\gamma$ since $(Rd_\star/d_U)^2\ge 4H^2B_\phi^2$ by~\eqref{eq:R_dstar}. Hence $\lambda_k\ge \gamma$ and~\eqref{eq:omega_bound} applies. We bound both of its terms by $(1-e^{-1})\sqrt{\lambda_k}/(2R)$. For the first, this is exactly~\eqref{eq:lambdan_warmup}:
\[
 2\sqrt{\frac{2d_\star\,\ell_{k,\delta}}{d_U\,m}} \le \frac{(1-e^{-1})\sqrt{\lambda_k}}{2R}.
\]
For the second, it holds whenever $\lambda_k\ge 16RHB_\phi\ell_{k,\delta}/((1-e^{-1})m)$, which~\eqref{eq:lambdan_warmup} implies by~\eqref{eq:R_dstar}. Consequently,
$\omega_k\le(1-e^{-1})\sqrt{\lambda_k}/R = (1-e^{-1})/R_k$, so the event $\cE_k$ holds, and Lemma~\ref{lem:localization} yields
\[
 \frac{R}{\sqrt{\lambda_k}}\,\norm{\t_k-\tE}_{\cH_k}\le1, \qquad \norm{\t_k-\tE}_{\cH_k}\le\frac{\omega_k}{1-e^{-1}}.
\]

\emph{Step 3: confidence bound.} Using~\eqref{eq:lambdan_warmup} and~\eqref{eq:R_dstar} again,
\[
 \frac{8HB_\phi\ell_{k,\delta}}{m\sqrt{\lambda_k}} \le \frac{1-e^{-1}}{\sqrt2} \sqrt{\frac{d_\star\,\ell_{k,\delta}}{d_U\,m}}, \quad\text{hence}\quad \omega_k \le \br{2\sqrt2+\frac{1-e^{-1}}{\sqrt2}}\sqrt{\frac{d_\star\,\ell_{k,\delta}}{d_U\,m}} \le 4\sqrt{\frac{d_\star\,\ell_{k,\delta}}{d_U\,m}}.
\]
Combining the last two bounds with~\eqref{eq:ell_bound} gives
\[
 \norm{\t_k-\tE}^2_{\cH_k} \le \frac{16\,d_\star\,\ell_{k,\delta}}{(1-e^{-1})^2 d_U m} \le \frac{16\,d_\star\log(3\eta d_\star km/\delta)}{(1-e^{-1})^2 m}.
\]
Every step holds simultaneously for all $n_\delta\le k\le n$ on the event from Lemma~\ref{lem:omega_concentration}, which has probability at least $1-\delta$.
\end{proof}

\section{Proofs of Section~\ref{sec:dynamics_selection}}
\label{apx:dynamics_selection}

We now lay out the results employed to prove Lemma~\ref{lemma:transfer}, Lemma~\ref{lemma:passive_learning}, Theorem~\ref{theo:active_learning} and Theorem~\ref{theorem:transfer}

\subsection{Supporting lemmas}

We first present useful identities of the suboptimality~\eqref{eq:objective}. 

\begin{proposition}[Remainder form]
\label{prop:subopt_remainder_form}
Let Assumption~\ref{as:linear_rewards} hold and let $P$ be any dynamics. For any $\t_1, \t_2 \in \R^d$,
\[
    V^\star_P(\t_2) - V^{\pi_P(\t_1)}_P(\t_2) = \norm{\t_2 - \t_1}_{G_P(\t_1, \t_2)}^2,
\]
where $G_P$ is defined in Lemma~\ref{lem:curvature_stability}.
\end{proposition}
\begin{proof}
Let $u \coloneq \t_2 - \t_1$. A Taylor expansion of $V^\star_P$ around $\t_1$ with integral remainder, together with Lemma~\ref{lemma:value_derivatives}, gives
\begin{align*}
    V^\star_P(\t_2) &= V^\star_P(\t_1) + DV^\star_P(\t_1)[u] + \int_0^1 (1-t)\,D^2V^\star_P(\t_1+tu)[u,u]\,\mathrm{d}t\\
    &= V^\star_P(\t_1) +\ip{u}{\phi_P(\t_1)} + \norm{u}_{G_P(\t_1, \t_2)}^2.
\end{align*}
On the other hand, the regularized value of the fixed policy $\pi_P(\t_1)$ is affine in the reward parameters, with slope $\phi_P(\t_1)$, and equals $V^\star_P(\t_1)$ at $\t_1$. Hence
\[
    V^{\pi_P(\t_1)}_P(\t_2) = V^\star_P(\t_1) + \ip{u}{\phi_P(\t_1)}.
\]
Subtracting the two expressions gives the result.
\end{proof}

\begin{proposition}[Pseudo self-concordant suboptimality bound]
\label{prop:value_gap_upper_bound}
Let Assumption~\ref{as:linear_rewards} hold, let $P$ be any dynamics, and let $\t_1, \t_2 \in \R^d$. Then, for any symmetric positive semidefinite matrix $X\in\R^{d\times d}$ with $\im\cH_P(\t_2)\subseteq\im X$,
\[
    V^\star_P(\t_2) - V^{\pi_P(\t_1)}_P(\t_2) \;\le\; \psi\br{R \norm{\t_2 - \t_1}_{\Pi_P}}\,\norm{X^{\dagger/2}\cH_P(\t_2) X^{\dagger/2}}_2\,\norm{\t_2 - \t_1}^2_{X},
\]
where $\psi$ is defined in Lemma~\ref{lem:curvature_stability}.
\end{proposition}
\begin{proof}
Set $u \coloneq \t_2-\t_1$ and $\psi_u \coloneq \psi\br{R \norm{u}_{\Pi_P}}$. By Proposition~\ref{prop:subopt_remainder_form} and item~2 of Lemma~\ref{lem:curvature_stability},
\[
    V^\star_P(\t_2) - V^{\pi_P(\t_1)}_P(\t_2) = \norm{u}_{G_P(\t_1, \t_2)}^2 \le \psi_u\,\norm{u}_{\cH_P(\t_2)}^2.
\]
Since $X^{1/2}X^{\dagger/2} = X^{\dagger/2}X^{1/2}$ is the orthogonal projector onto $\im X$, and $\im\cH_P(\t_2)\subseteq\im X$, we have $\cH_P(\t_2) = X^{1/2}\br{X^{\dagger/2}\cH_P(\t_2)X^{\dagger/2}}X^{1/2}$. Hence
\[
    \norm{u}_{\cH_P(\t_2)}^2
    = \br{X^{1/2}u}^\top\br{X^{\dagger/2}\cH_P(\t_2)X^{\dagger/2}}\br{X^{1/2}u}
    \le \norm{X^{\dagger/2}\cH_{P}(\t_2) X^{\dagger/2}}_2\,\norm{u}_{X}^2,
\]
which concludes the proof.
\end{proof}

We now show that once the warm-up phase has terminated, i.e, condition~\eqref{eq:n_min}, pseudo self-concordance factors (Lemma~\ref{lem:self_conc}) are upper bounded by their localized constants.

\begin{lemma}[Localized self-concordance constants]
\label{lemma:local_self_concordance_constants}
Let Assumption~\ref{as:linear_rewards} hold and $\tE\in\T$. On the event of Lemma~\ref{lem:omega_concentration}, for every round $n_\delta\le k\le n$, where $n_\delta$ is defined by the warm-up condition~\eqref{eq:n_min}, and every $P \in \cP$,
\[
\psi\br{R \norm{\tE - \t_k}_{\Pi_P}} \le 1, \qquad \exp\br{R\norm{\tE - \t_k}_{\Pi_P}} \le e.
\]
\end{lemma}
\begin{proof}
Fix $P\in \cP$ and $n_\delta\le k\le n$. By Step~1 of the proof of Lemma~\ref{thm:confidencenew}, $\lambda_k \ge \lambda_{n_\delta} > 0$. Since $U_P\subseteq U$ and $\cH_k\succeq\lambda_k\Pi_U$ (see the proof of Lemma~\ref{lemma:loss_self_conf}),
\[
    R\norm{\tE - \t_k}_{\Pi_P} \le R\norm{\tE - \t_k}_{\Pi_U} \le R\,\frac{\norm{\tE-\t_k}_{\cH_k}}{\sqrt{\lambda_k}} = R_k\,\rho_k.
\]
By Step~2 of the proof of Lemma~\ref{thm:confidencenew}, the localization event $\cE_k$~\eqref{eq:local_event} holds, so Lemma~\ref{lem:localization} gives $R_k\rho_k\le1$. Since $\psi$ and $\exp$ are increasing,
\[
\psi\br{R \norm{\tE - \t_k}_{\Pi_P}} \le \psi(1) = 1
\qquad\text{and}\qquad
\exp\br{R \norm{\tE - \t_k}_{\Pi_P}} \le e.
\]
\end{proof}

Additionally, we provide a characterization of the lower bound of optimal value Hessians evaluated at the expert parameter $\tE$, as well as the resulting once-cycle coverage $\lambda_{\abs \cP}$ obtained after following round-robin for $\abs \cP$ rounds.

\begin{lemma}[Hessian lower bound]
\label{lemma:Hessian_lb}
Assume finite $\cS$ and $\cA$, and $U\neq\{0\}$. Let $\nu^\text{E}_{h,P}(s)$ denote the probability that the expert visits $s$ at timestep $h$ under $P$, and call $(h,s)$ \emph{reachable} under $P$ when this probability is positive. For $v\in\R^d$, let $\alpha^v_{h,P}(s,a) \coloneq \ip{v}{A^{\tE}_{h,P}(s,a)}$, and define the minimal expert occupancy and the minimal identifiable advantage gap
\[
    \mu_{\min} \coloneq \min_{\substack{P\in\cP,\ a\in\cA \\ (h,s)\ \text{reachable}}} \nu^\text{E}_{h,P}(s)\,\piE_{h,P}(a\mid s),
    \qquad
    \Delta \coloneq \min_{\substack{P\in\cP:\ U_P\ne\{0\} \\ v\in U_P,\ \norm{v}_2=1}}\ \max_{\substack{a \in \cA \\ (h,s)\ \text{reachable}}} \abs{\alpha^v_{h,P}(s, a)} .
\]
Then $\mu_{\min}>0$, $\Delta>0$, and for every $P\in\cP$,
\[
    \cH_P(\tE) \;\succeq\; \kappa\, \Pi_P,
    \qquad \kappa \coloneq \eta\,\mu_{\min}\,\Delta^2.
\]
\end{lemma}
\begin{proof}
Fix $P\in\cP$ and $v\in\R^d$.

\emph{Step 1 (per-step decomposition).} By definition of the feature advantage, $\mathbb{E}_{a\sim\piE_{h,P}(\cdot\mid s)}\bs{\alpha^v_{h,P}(s,a)} = 0$ for every $(h,s)$. Hence $\br{\alpha^v_{h,P}(s_h,a_h)}_{h=1}^H$ is a martingale difference sequence along the expert trajectory, and its terms are uncorrelated. Since $\piE_P=\pi_P(\tE)$, Lemma~\ref{lemma:value_derivatives} gives
\[
    \norm{v}^2_{\cH_P(\tE)}
    = \eta\,\mathbb{E}\bs{\Big(\sum\nolimits_{h=1}^H \alpha^v_{h,P}(s_h,a_h)\Big)^2}
    = \eta \sum_{h=1}^H \sum_{s,a} \nu^\text{E}_{h,P}(s)\,\piE_{h,P}(a\mid s)\, \alpha^v_{h,P}(s,a)^2 .
\]

\emph{Step 2 (positivity of $\mu_{\min}$ and $\Delta$).} Since $\piE_{h,P}$ has full support~\eqref{eq:softmax_pi} and $\cS$, $\cA$ and $\cP$ are finite, $\mu_{\min}>0$.

By Step 1, $\norm{v}^2_{\cH_P(\tE)}$ is a sum of non-negative terms with positive weights on reachable pairs. Hence $v \in \ker\br{\cH_P(\tE)}$ if and only if $\alpha^v_{h,P}(s,a)=0$ for every $h$, every $s$ reachable at $h$, and every $a$. Since $U_P = \ker\br{\cH_P(\tE)}^\perp$ (Proposition~\ref{lemma:image}), every unit $v\in U_P$ lies outside the kernel, so the inner maximum defining $\Delta$ is strictly positive. This maximum is a maximum of finitely many continuous functions of $v$, hence continuous, and the unit sphere of $U_P$ is compact. The minimum over $v$ is therefore attained and strictly positive, and finiteness of $\cP$ gives $\Delta>0$.

\emph{Step 3 (conclusion).} If $U_P=\{0\}$, then $\cH_P(\tE)=0=\Pi_P$ and the claim holds trivially. Otherwise, let $w\in U_P$ be a unit vector, and let $(h^\star,s^\star,a^\star)$ attain the inner maximum in the definition of $\Delta$ for $w$. Keeping only that term in Step 1,
\[
    \norm{w}^2_{\cH_P(\tE)}
    \ \ge\ \eta\, \nu^\text{E}_{h^\star,P}(s^\star)\, \piE_{h^\star,P}(a^\star\mid s^\star)\, \alpha^w_{h^\star,P}(s^\star,a^\star)^2
    \ \ge\ \eta\, \mu_{\min}\, \Delta^2 = \kappa .
\]
For general $v\in\R^d$, since $\cH_P(\tE)$ vanishes on $U_P^\perp$, $\norm{v}^2_{\cH_P(\tE)} = \norm{\Pi_Pv}^2_{\cH_P(\tE)} \ge \kappa\norm{\Pi_Pv}_2^2 = \kappa\, v^\top\Pi_Pv$, which is $\cH_P(\tE)\succeq\kappa\,\Pi_P$.
\end{proof}

\begin{corollary}[Information coverage after one cycle]\label{cor:roundrobin_coverage}
Assume $U\ne\{0\}$. After selecting each dynamics $P\in\cP$ once,
\[
    \lambda_{\abs{\cP}} = \lambda_{\min}^{U}\!\left( \sum_{P\in\cP}\cH_P(\tE)\right)\ge\eta\mu_{\min}\Delta^2\min_{\substack{v\in U\\ \norm{v}_2=1}}\sum_{P\in\cP}\norm{\Pi_Pv}_2^2>0,
\]
where $\mu_{\min}$ and $\Delta$ are defined in Lemma~\ref{lemma:Hessian_lb}.
\end{corollary}

\begin{proof}
After one complete cycle, $\cH_{\abs{\cP}}=\sum_{P\in\cP}\cH_P(\tE)$. By Lemma~\ref{lemma:Hessian_lb},
\[
    \cH_{\abs{\cP}}\succeq\eta\mu_{\min}\Delta^2\sum_{P\in\cP}\Pi_P,
\]
and since $v^\top\Pi_Pv=\norm{\Pi_Pv}_2^2$, taking the minimum of the quadratic form over unit vectors in $U$ gives the lower bound. Moreover, since each $\Pi_P$ is an orthogonal projector and $U=\sum_{P\in\cP}U_P$,
\[
    \ker\!\left(\sum_{P\in\cP}\Pi_P\right) = \bigcap_{P\in\cP}U_P^\perp = U^\perp,
\]
so $\sum_{P\in\cP}\Pi_P$ is positive definite on $U$, and its minimum quadratic form over the unit sphere of $U$ is strictly positive.
\end{proof}

\begin{proposition} 
\label{prop:linear_vs_log}
Let $a, b > 0$ with $ab \ge 1$. Then $x \ge a\log(bx)$ for every $x \ge 2a\log(2ab)$.
\end{proposition}
\begin{proof}
The map $f(x) \coloneq x - a\log(bx)$ is increasing for $x \ge a$, and $x_0 \coloneq 2a\log(2ab) \ge a$ since $ab \ge 1$. Using $\log y \le y$, we get $f(x_0) = a\br{\log(2ab) - \log\log(2ab)} \ge 0$.
\end{proof}

\transferability*
\begin{proof}
We work on the event of Lemma~\ref{lem:omega_concentration}, which has probability at least $1-\delta$. Fix $P\in\cP$ and a round $n_\delta\le k\le n$, and let $u_k \coloneq \t_k-\tE$. Recall that
\[
    \subopt(\t_k,P) = V^\star_P(\tE) - V^{\pi_P(\t_k)}_P(\tE),
    \qquad
    g_k(P) = \norm{\cH_k^{\dagger/2}\,\cH_P(\tE)\,\cH_k^{\dagger/2}}_2.
\]

\emph{Step 1 (image condition).} By Step~1 of the proof of Lemma~\ref{thm:confidencenew}, $\lambda_k\ge\lambda_{n_\delta}>0$. Since $\cH_k$ vanishes on $U^\perp$ and $\cH_k\succeq\lambda_k\Pi_U$, we have $\im\cH_k = U$. As $\im\cH_P(\tE) = U_P\subseteq U$ by Proposition~\ref{lemma:image}, it follows that $\im\cH_P(\tE)\subseteq\im\cH_k$.

\emph{Step 2 (change of geometry).} By Step~1, Proposition~\ref{prop:value_gap_upper_bound} applies with $\t_1 = \t_k$, $\t_2 = \tE$, and $X = \cH_k$:
\[
    \subopt(\t_k,P)
    \;\le\; \psi\br{R\norm{\tE-\t_k}_{\Pi_P}}\,
    \norm{\cH_k^{\dagger/2}\,\cH_P(\tE)\,\cH_k^{\dagger/2}}_2\,
    \norm{u_k}^2_{\cH_k}
    \;=\; \psi\br{R\norm{\tE-\t_k}_{\Pi_P}}\, g_k(P)\,\norm{u_k}^2_{\cH_k}.
\]

\emph{Step 3 (localization).} By Lemma~\ref{lemma:local_self_concordance_constants}, $\psi\br{R\norm{\tE-\t_k}_{\Pi_P}}\le1$, hence
\[
    \subopt(\t_k,P) \;\le\; g_k(P)\,\norm{u_k}^2_{\cH_k}.
\]

\emph{Step 4 (confidence bound).} By Lemma~\ref{thm:confidencenew}, $\norm{u_k}^2_{\cH_k}\le\beta_k^2(\delta)$, which gives
\[
    \subopt(\t_k,P)\le g_k(P)\,\beta^2_k(\delta).
\]
All steps hold on the same event, simultaneously for every $P\in\cP$ and $n_\delta\le k\le n$, which concludes the proof.
\end{proof}

\roundrobin*
\begin{proof}
By definition of $\beta_n(\delta)$ in Lemma~\ref{thm:confidencenew}, the warm-up condition~\eqref{eq:n_min} reads
\[
    \lambda_k \;\ge\; 2R^2\beta_n^2(\delta).
\]
Fix a round $k\le n$ and let $j \coloneq \floor{k/\abs\cP}$ be the number of completed round-robin cycles. Each $P\in\cP$ has been selected at least $j$ times, so $\cH_k \succeq j\,\cH_{\abs\cP}$ and hence $\lambda_k \ge j\,\lambda_{\abs\cP}$, where $\lambda_{\abs\cP}>0$ by Corollary~\ref{cor:roundrobin_coverage}. The warm-up condition therefore holds at round $k$ whenever
\[
    j \;\ge\; \frac{2R^2\beta_n^2(\delta)}{\lambda_{\abs\cP}},
\]
which is the case for $k = \abs\cP\roof{2R^2\beta_n^2(\delta)/\lambda_{\abs\cP}}$. If this round is at most $n$, the definition of $n_\delta$ as the first round satisfying~\eqref{eq:n_min} gives the claim.
\end{proof}

\section{Proofs of section~\ref{sec:transfer}}
\label{apx:transfer}
Having stated the required lemmas, we subsequently prove our main transfer results. We first present the general transfer guarantee.

\greedy*
\begin{proof}
\emph{Setup.}
We work on the event of Lemma~\ref{thm:confidencenew}, which holds with probability at least $1-\delta$, simultaneously for all rounds. Let $\Lambda\coloneq 32R^2d_\star/((1-e^{-1})^2m)$, which lower-bounds the right-hand side of~\eqref{eq:n_min} since $3\eta d_\star nm/\delta\ge e$. By definition of $n_\delta$, every round $k\ge n_\delta$ satisfies $\lambda_k\ge\Lambda>0$, so $\cH_k$ is invertible on $U$, and the confidence and localization bounds apply. The same holds for the plug-in matrix $\hat\cH_k\coloneq\sum_{i=1}^k\cH_{P_i}(\t_k)$, which has the same image $U$ since Hessian images do not depend on the parameters (Proposition~\ref{lemma:image}). For $k\ge n_\delta$, we write
\[
    \cT_k(P) \coloneq \cH_k^{\dagger/2}\,\cH_P(\tE)\,\cH_k^{\dagger/2},
    \qquad
    \hat\cT_k(P) \coloneq \hat\cH_k^{\dagger/2}\,\cH_P(\t_k)\,\hat\cH_k^{\dagger/2},
\]
so that $g_k(P)=\norm{\cT_k(P)}_2$ and $\hat g_k(P)=\norm{\hat\cT_k(P)}_2$. All matrices below are restricted to $U$. Since $\norm{\cH_P(\tE)}_2\le\eta H^2B_\phi^2$ (Proposition~\ref{proposition:hessian_ub}) and $\cH_k\succeq\Lambda I$ on $U$,
\begin{equation}
\label{eq:greedy_bounded}
    \cT_k(P) \;\preceq\; c_m\, I,
    \qquad
    c_m \coloneq \frac{\eta H^2B_\phi^2}{\Lambda} = \frac{(1-e^{-1})^2\,m}{32\,(H+1)^2\,\eta d_\star},
\end{equation}
where we used $R^2=\eta^2H^2(H+1)^2B_\phi^2$.

\emph{Step 1 (localization of $g_k$).}
Fix $k\ge n_\delta$. By the proof of Lemma~\ref{lemma:local_self_concordance_constants}, $R\norm{\t_k-\tE}_{\Pi_P}\le1$ for every $P\in\cP$, so item~1 of Lemma~\ref{lem:curvature_stability} gives $e^{-1}\cH_P(\tE)\preceq\cH_P(\t_k)\preceq e\,\cH_P(\tE)$. Summing over the selected dynamics, $e^{-1}\cH_k\preceq\hat\cH_k\preceq e\,\cH_k$.

For positive definite $B$ and positive semidefinite $A$ on $U$, the Courant--Fischer theorem gives that the $i$-th largest eigenvalue of $B^{-1/2}AB^{-1/2}$ equals $\max_{\dim\cV = i}\ \min_{v\in\cV\setminus\{0\}} \norm{v}^2_A/\norm{v}^2_B$, and each such ratio increases with $A$ and decreases with $B$ in the Loewner order. Hence every eigenvalue of $\hat\cT_k(P)$ lies within a factor $e^2$ of the corresponding eigenvalue of $\cT_k(P)$, and in particular
\begin{equation}
\label{eq:greedy_plugin}
    e^{-2}\,g_k(P) \;\le\; \hat g_k(P) \;\le\; e^{2}\,g_k(P) \qquad\text{for every } P\in\cP.
\end{equation}

\emph{Step 2 (greedy rule)}
For $k\ge n_\delta$, the rule selects $P_{k+1}\in\argmax_{P\in\cP}\hat g_k(P)$. By~\eqref{eq:greedy_plugin}, for every $P\in\cP$,
\[
    g_k(P) \;\le\; e^{2}\,\hat g_k(P) \;\le\; e^{2}\,\hat g_k(P_{k+1}) \;\le\; e^{4}\,g_k(P_{k+1}).
\]
Since $\log(1+x)\ge x/(1+x)$, every $x\in[0,c_m]$ satisfies $x\le\bar c_m\log(1+x)$ with $\bar c_m\coloneq 1+c_m$. Applying this to the largest eigenvalue of $\cT_k(P_{k+1})$, which lies in $[0,c_m]$ by~\eqref{eq:greedy_bounded}, and using that the other eigenvalues contribute non-negatively, gives
\begin{equation}
\label{eq:greedy_selection}
    g_k(P) \;\le\; e^{4}\,\bar c_m\,\log\det\br{I+\cT_k(P_{k+1})}
    \qquad\text{for every } P\in\cP.
\end{equation}

\emph{Step 3 (elliptical potential lemma).}
Fix $P\in\cP$ and $n_\delta<k\le n$. Since $\cH_k\succeq\cH_i$ on $U$ for $i\le k$, we have $g_k(P)\le g_i(P)$. Averaging over $i\in\{n_\delta,\dots,k-1\}$ and applying~\eqref{eq:greedy_selection},
\[
    g_k(P)
    \;\le\; \frac{e^{4}\,\bar c_m}{k-n_\delta}\sum_{i=n_\delta}^{k-1}\log\det\br{I+\cT_i(P_{i+1})}.
\]
Since $\cH_{i+1} = \cH_i + \cH_{P_{i+1}}(\tE)$, we have $\det\cH_{i+1}=\det\cH_i\,\det\br{I+\cT_i(P_{i+1})}$ on $U$, so the sum telescopes to $\log\br{\det\cH_k/\det\cH_{n_\delta}}$. On $U$, every eigenvalue of $\cH_k$ is at most $k\eta H^2B_\phi^2$ (Proposition~\ref{proposition:hessian_ub}) and every eigenvalue of $\cH_{n_\delta}$ is at least $\Lambda$, so every eigenvalue of $\cH_{n_\delta}^{-1/2}\cH_k\cH_{n_\delta}^{-1/2}$ is at most $k\,c_m$. Since $U$ has dimension $d_U$, $\log\br{\det\cH_k/\det\cH_{n_\delta}} \le d_U\log(1+k\,c_m)$. Taking the maximum over $P\in\cP$,
\begin{equation}
\label{eq:greedy_gn}
    \max_{P\in\cP}\, g_k(P) \;\le\; e^{4}\,\bar c_m\,\frac{d_U\log(1+k\,c_m)}{k-n_\delta}.
\end{equation}
Since $\eta d_\star\ge1$, we have $c_m\le m$ and $c_m\lesssim m/(\eta d_\star H^2)$, and $\log(1+k\,c_m)\le\log(1+km)\le2\log(km)$ as $km\ge2$. This proves the first claim.

\emph{Step 4 (suboptimality bound).}
By Lemma~\ref{lemma:transfer} and~\eqref{eq:greedy_gn} at $k=n$, and since $\beta_n^2(\delta)\lesssim d_\star\log(\eta d_\star nm/\delta)/m$ and $c_m\,d_\star/m\lesssim1/(\eta(H+1)^2)$, it holds for every $P\in\cP$ that
\[
    \subopt(\t_n,P)
    \;\le\; g_n(P)\,\beta_n^2(\delta)
    \;\lesssim\; \br{\frac{d_\star}{m}+\frac{1}{\eta(H+1)^2}}\frac{d_U\log(nm)\log(\eta d_\star nm/\delta)}{n-n_\delta}.
\]
If $m\le\eta d_\star H^2$, then $\frac{1}{\eta(H+1)^2}\le\frac{1}{\eta H^2}\le\frac{d_\star}{m}$, and if $n\ge2n_\delta$, then $n-n_\delta\ge n/2$. Together, these give~\eqref{eq:main_rate}, on the same event.
\end{proof}

\begin{remark}[Alternative selection criteria]
\label{rem:criteria_proof}
Recall that the active rule~\eqref{eq:active_selection_rule} maximizes $\hat g_k(P)=\norm{\hat\cT_k(P)}_2$, where
\[
    \hat\cT_k(P) \coloneq \hat\cH_k^{\dagger/2}\,\cH_P(\t_k)\,\hat\cH_k^{\dagger/2},
    \qquad
    \hat\cH_k \coloneq \sum_{j=1}^k\cH_{P_j}(\t_k),
\]
is the plug-in estimate of $\cT_k(P) = \cH_k^{\dagger/2}\,\cH_P(\tE)\,\cH_k^{\dagger/2}$. Instead of the spectral norm, the rule may maximize the trace $\tr\hat\cT_k(P)$ or the log-determinant $\log\det(I+\hat\cT_k(P))$. Unlike the spectral norm, these criteria account for all directions that matter under $P$, not only the least covered one; the log-determinant is the variant used in our experiments (Section~\ref{sec:experiments}). Theorem~\ref{theo:active_learning} holds for both.

Indeed, Steps~3 and~4 of the proof above only use~\eqref{eq:greedy_selection}, so it suffices to show that it still holds for these choices of $P_{k+1}$. By Step~1, every eigenvalue of $\hat\cT_k(P)$ lies within a factor $e^2$ of the corresponding eigenvalue of $\cT_k(P)$.

\emph{Trace.} Comparing eigenvalues and using the choice of $P_{k+1}$,
\[
    g_k(P)\;\le\;\tr\cT_k(P)\;\le\;e^2\tr\hat\cT_k(P)\;\le\;e^2\tr\hat\cT_k(P_{k+1})\;\le\;e^4\tr\cT_k(P_{k+1}).
\]
Applying the scalar bound $x\le\bar c_m\log(1+x)$ of Step~2 to each eigenvalue of $\cT_k(P_{k+1})$ gives $\tr\cT_k(P_{k+1})\le\bar c_m\log\det\br{I+\cT_k(P_{k+1})}$.

\emph{Log-determinant.} By the same scalar bound, $g_k(P)\le\bar c_m\log\det\br{I+\cT_k(P)}$. Since $\log(1+cx)\le c\log(1+x)$ for $c\ge1$, comparing eigenvalues as above gives
\[
    \log\det\br{I+\cT_k(P)}\;\le\;e^2\log\det\br{I+\hat\cT_k(P)}\;\le\;e^2\log\det\br{I+\hat\cT_k(P_{k+1})}\;\le\;e^4\log\det\br{I+\cT_k(P_{k+1})}.
\]
In both cases,~\eqref{eq:greedy_selection} holds, and the theorem follows identically.
\end{remark}

We now provide the proof of Theorem~\ref{theorem:transfer} which allows us to transfer the recovered reward to arbitrary unseen dynamics. We introduce a couple of quantities required for our analysis. We measure how $P$ relates to the training set $\cP$ through the following quantities:
\begin{align*}
g_n(\cP) &:= \max_{P\in\cP} g_n(P), \quad
s_P := \norm{\Pi_P\Pi_{U^\perp}}_2, \quad
\gamma(P) := \max_{w\in U\setminus\{0\}}
\frac{\norm{w}_{\cH_P(\tE)}^2}{\max_{Q\in\cP}\norm{w}_{\cH_Q(\tE)}^2}.
\end{align*}
Here, $s_P\in[0,1]$ measures how far the identifiable subspace of $P$ leaves $U$, and $\gamma(P)$ compares the curvature of $P$ along $U$ with that of the training dynamics. The ratio defining $\gamma(P)$ is well defined. If $w\in U$ satisfies $\norm{w}_{\cH_Q(\tE)}=0$ for every $Q\in\cP$, then $w\in\bigcap_{Q\in\cP}U_Q^\perp=U^\perp$ by Proposition~\ref{lemma:image}, so $w=0$. By homogeneity, the maximum may be taken over the unit sphere of $U$, hence it is attained and finite. Moreover, $\gamma(P)\le1$ for every $P\in\cP$, since $P$ itself appears in the inner maximum. 

Note that $s_P=0$ and $g_\perp(P)=0$ are both equivalent to $U_P\subseteq U$.

\transferabilitybeyond*

\begin{proof}
We work on the confidence event of Lemma~\ref{thm:confidencenew}. Since $n\ge2n_\delta\ge n_\delta$, its bounds and those of Lemma~\ref{lemma:local_self_concordance_constants} hold at round $n$. Let
\begin{equation*}
\Delta:=\theta_n-\tE=u+v,
\qquad
u:=\Pi_U\Delta,
\qquad
v:=\Pi_{U^\perp}\Delta.
\end{equation*}
Since $\cH_n$ vanishes on $U^\perp$ and $\cH_n\succeq\lambda_n\Pi_U$,
the confidence bound, the localization of Lemma~\ref{lemma:local_self_concordance_constants}, and
boundedness of $\Theta$ give
\begin{equation*}
\norm{u}_{\cH_n}\le\beta_n(\delta),
\qquad
R\norm{u}_2\le1,
\qquad
\norm{v}_2\le2B_\theta.
\end{equation*}
Since $\operatorname{im}\cH_Q(\tE)\subseteq U=\operatorname{im}\cH_n$ for every $Q\in\cP$, the proof of Lemma~\ref{lemma:transfer} gives $\norm{u}_{\cH_Q(\tE)}^2\le g_n(Q)\norm{u}_{\cH_n}^2$.
Thus, by definition of $\gamma(P)$ and since $u\in U$, it follows that
\begin{align*}
\norm{u}_{\cH_P(\tE)}^2
&\le
\gamma(P)\max_{Q\in\cP}
\norm{u}_{\cH_Q(\tE)}^2\\
&\le
\gamma(P)g_n(\cP)\norm{u}_{\cH_n}^2\\
&\le
\gamma(P)g_n(\cP)\beta_n^2(\delta).
\end{align*}
Moreover, $v=\Pi_{U^\perp}v$ implies
\begin{align*}
\norm{v}_{\cH_P(\tE)}^2
&=
v^\top\Pi_{U^\perp}\cH_P(\tE)\Pi_{U^\perp}v\\
&\le g_\perp(P)\norm{v}_2^2\\
&\le 4g_\perp(P)B_\theta^2.
\end{align*}
Consequently,
\begin{align*}
\norm{\Delta}_{\cH_P(\tE)}^2
&\le
2\norm{u}_{\cH_P(\tE)}^2
+2\norm{v}_{\cH_P(\tE)}^2\\
&\le
2\gamma(P)g_n(\cP)\beta_n^2(\delta)
+8g_\perp(P)B_\theta^2.
\end{align*}

Furthermore, $v=\Pi_{U^\perp}v$ and the definition of $s_P$ yield
\begin{align*}
R\norm{\Pi_P\Delta}_2
&\le R\norm{u}_2+R\norm{\Pi_P\Pi_{U^\perp}v}_2\\
&\le
R\norm{u}_2+R\,s_P\norm{v}_2\\
&\le
1+2RB_\theta s_P.
\end{align*}
Applying Proposition~\ref{prop:subopt_remainder_form} and item~2 of Lemma~\ref{lem:curvature_stability} with $\theta_1=\theta_n$ and $\theta_2=\tE$, the monotonicity of $\psi$, and $\psi(1+a)\le e\psi(a)$
for $a\ge0$, which follows from $\psi(x)=\int_0^1te^{xt}\,dt$, yields
\begin{align}
\subopt(\theta_n,P)
&\le
\psi\!\left(R\norm{\Pi_P\Delta}_2\right)
\norm{\Delta}_{\cH_P(\tE)}^2 \nonumber\\
&\le
\psi\!\left(1+2RB_\theta s_P\right)
\norm{\Delta}_{\cH_P(\tE)}^2 \nonumber\\
&\le
2e\psi(2RB_\theta s_P)\Bigl[
\gamma(P)g_n(\cP)\beta_n^2(\delta)
+4g_\perp(P)B_\theta^2
\Bigr]. \label{eq:transfer_explicit}
\end{align}
Since $m\le\eta d_\star H^2$ and $n\ge2n_\delta$, Theorem~\ref{theo:active_learning} and the definition of $\beta_n(\delta)$ give
\begin{equation*}
g_n(\cP)\beta_n^2(\delta)
\lesssim
\frac{d_\star d_U\log(nm)\log(\eta d_\star nm/\delta)}{nm},
\end{equation*}
which, plugged into~\eqref{eq:transfer_explicit}, yields the claim. The confidence event does not depend on $P$, so the bound holds simultaneously for all $P$.

Finally, $U_P\subseteq U$ implies $s_P=g_\perp(P)=0$, so the transfer bias vanishes, and $\psi(0)=1/2$.
For $P\in\cP$, the definition of $\gamma(P)$
directly gives $\gamma(P)\le1$, and we recover Theorem~\ref{theo:active_learning}.
\end{proof}

\begin{remark}[Connection to principal angles]
Suppose $\dim U_P\le\dim U$ and denote
$\alpha_P:=\alpha_{\max}(U_P,U)$ for the maximal principal angle between $U_P$ and $U$.
Then
\begin{equation*}
s_P=\norm{\Pi_P\Pi_{U^\perp}}_2=\sin\alpha_P,
\end{equation*}
so the constant in~\eqref{eq:transfer_explicit} reads $\psi(2RB_\theta\sin\alpha_P)$.
In particular, since $\psi(1)=1$, it is at most one whenever
$2RB_\theta\sin\alpha_P\le1$. If instead $\dim U_P>\dim U$, then $U_P\cap U^\perp\neq\{0\}$ and $s_P=1$.

Moreover, since $\cH_P(\tE)^{1/2}=\cH_P(\tE)^{1/2}\Pi_P$,
\begin{align*}
g_\perp(P)
&=\norm{\cH_P(\tE)^{1/2}\Pi_P\Pi_{U^\perp}}_2^2\\
&\le \norm{\cH_P(\tE)}_2\sin^2\alpha_P.
\end{align*}
Both the constant and the transfer bias are thus controlled by $\alpha_P$.
For a single training dynamics $P^0$ with
$\dim U_{P^0}=\dim U_P$, we have
\begin{equation*}
\alpha_P
=\alpha_{\max}(U_{P^0}^\perp,U_P^\perp).
\end{equation*}
This is the angle between unidentifiable subspaces
appearing in the transferability guarantee of~\citet[Theorem~3.11]{schlaginhaufen2024towards}.
\end{remark}

\section{Offline Active Learning}
\label{app:offline_active_learning}

In some cases, sequential interaction with the expert may be impractical, for instance due to limited expert availability or high environment setup costs. We therefore also consider an offline variant, where the entire sequence of dynamics is selected in advance. Setting $\t_0 \coloneq 0$ and $\varepsilon\coloneq\eta H^2B_\phi^2$, we construct the offline information proxy and offline transfer operator
\[
    \cH^0_k \coloneq \varepsilon\,\Pi_U + \sum_{i=1}^{k} \cH_{P_i}(\t_0),
    \qquad
    \cT^0_k(P) \coloneq (\cH^0_k)^{\dagger/2}\,\cH_P(\t_0)\,(\cH^0_k)^{\dagger/2},
\]
and select dynamics with
\begin{equation}
\label{eq:active_selection_rule_offline}
P_{k+1} \in \argmax_{P \in \cP}\, J \br{\cT^0_k(P)}, \quad \text{with} \quad J \in \bc{\norm{\cdot}_2,\, \tr,\, \log\det(I + \cdot)}.
\end{equation}
Since $U$ and $\cH_P(\t_0)$ do not depend on expert demonstrations, this rule constructs the full sequence $(P_k)_{k=1}^n$ without any interaction with the expert.

\begin{theorem}
\label{lemma:offline_transfer_bound}
Let Assumption~\ref{as:linear_rewards} hold, assume finite $\cS$ and $\cA$ and $U\neq\{0\}$, and follow~\eqref{eq:active_selection_rule_offline} at every round. Let $\kappa$ be defined in Lemma~\ref{lemma:Hessian_lb}, and $\xi\coloneq\min_{v\in U,\norm{v}_2=1}\max_{P\in\cP}\norm{v}_{\Pi_P}^2>0$. Then, for every $n\ge n_\delta$ and $P\in\cP$,
\[
    g_n(P) \;\le\; 2e^{2RB_\t}\br{1+\frac{m}{\eta d_\star H^2}}\frac{d_U\log(1+n)}{n},
\]
and $n_\delta\le\roof{2K\log(4K)}$ whenever the right-hand side is at most $n$, where
\[
    K\coloneq\frac{2d_U\,e^{2RB_\t}\br{2R^2\beta_n^2(\delta)+\eta H^2B_\phi^2}}{\kappa\,\xi}.
\]
\end{theorem}

\begin{proof}
All matrices below are positive semidefinite and vanish on $U^\perp$, so we compare them in the Loewner order on $U$, where $\cH^0_k\succeq\varepsilon\Pi_U$ is invertible and $(\cH^0_k)^{\dagger}$ is its inverse. Determinants $\det\nolimits_U$ are taken on $U$, of dimension $d_U$. Let $t_k\coloneq\max_{P\in\cP}\norm{\cT^0_k(P)}_2$ and $a\coloneq RB_\t$.

\emph{Step 1 (elliptical potential lemma).} Since $\cH^0_k\succeq\varepsilon\,\Pi_U$ and $\cH_P(\t_0)\preceq\eta H^2B_\phi^2\,\Pi_U$ (Proposition~\ref{proposition:hessian_ub}), every eigenvalue of $\cT^0_k(P)$ lies in $[0,1]$. As $x\le2\log(1+x)$ on $[0,1]$, for every $P\in\cP$ and each criterion $J$,
\[
    \norm{\cT^0_k(P)}_2 \;\le\; 2\log\det\br{I+\cT^0_k(P_{k+1})}.
\]
For $J=\log\det(I+\cdot)$, this follows from $\norm{\cT^0_k(P)}_2\le2\log\det(I+\cT^0_k(P))$ and the choice of $P_{k+1}$. For $J=\norm{\cdot}_2$ and $J=\tr$, it follows from $\norm{\cT^0_k(P)}_2\le J(\cT^0_k(P))\le J(\cT^0_k(P_{k+1}))$ and the scalar bound applied to each eigenvalue of $\cT^0_k(P_{k+1})$. Since $\cH^0_k$ is nondecreasing in $k$, $\norm{\cT^0_n(P)}_2\le\norm{\cT^0_k(P)}_2$ for $k\le n$. Averaging over $k\in\{0,\dots,n-1\}$ and using $\det\nolimits_U\cH^0_{k+1}=\det\nolimits_U\cH^0_k\,\det\br{I+\cT^0_k(P_{k+1})}$,
\[
    t_n \;\le\; \frac{2}{n}\sum_{k=0}^{n-1}\log\det\br{I+\cT^0_k(P_{k+1})}
    \;=\; \frac{2}{n}\log\frac{\det\nolimits_U\cH^0_n}{\varepsilon^{d_U}}
    \;\le\; \frac{2d_U\log(1+n)}{n},
\]
where the last step uses that every eigenvalue of $\cH^0_n$ on $U$ is at most $\varepsilon+n\eta H^2B_\phi^2=(1+n)\varepsilon$.

\emph{Step 2 (from $\t_0$ to $\tE$).} Since $\norm{\tE}_{\Pi_P}\le\norm{\tE}_2\le B_\t$ for every $P\in\cP$, item~1 of Lemma~\ref{lem:curvature_stability} with $\t_1=\tE$ and $\t_2=\t_0$ gives $e^{-a}\cH_P(\t_0)\preceq\cH_P(\tE)\preceq e^{a}\cH_P(\t_0)$. Summing over the selected dynamics,
\begin{equation}
\label{eq:offline_proxy_comparison}
    \cH^0_k \;\preceq\; \varepsilon\,\Pi_U + e^{a}\,\cH_k.
\end{equation}

\emph{Step 3 (transfer bound).} Let $n\ge n_\delta$, so that $\lambda_n\ge\Lambda$, with $\Lambda$ as in the proof of Theorem~\ref{theo:active_learning}, and $\varepsilon\Pi_U\preceq(\varepsilon/\Lambda)\cH_n = c_m\cH_n$. By definition of $t_n$, Step~2 and~\eqref{eq:offline_proxy_comparison},
\[
    \cH_P(\tE)\;\preceq\;e^{a}\cH_P(\t_0)\;\preceq\;e^{a}\,t_n\,\cH^0_n\;\preceq\;e^{a}\,t_n\br{c_m+e^{a}}\cH_n,
\]
hence $g_n(P)=\norm{\cH_n^{\dagger/2}\cH_P(\tE)\cH_n^{\dagger/2}}_2\le e^{a}(e^{a}+c_m)\,t_n\le e^{2a}(1+c_m)\,t_n$. By~\eqref{eq:greedy_bounded}, $c_m\le m/(\eta d_\star H^2)$, and Step~1 gives the first claim.

\emph{Step 4 (offline warm-up).} Let $v\in U$ be a unit vector and $P^\star\in\argmax_{P\in\cP}\norm{\Pi_Pv}_2^2$, so that $\norm{\Pi_{P^\star}v}_2^2\ge\xi$. By Lemma~\ref{lemma:Hessian_lb} and Step~2, $\cH_{P^\star}(\t_0)\succeq e^{-a}\cH_{P^\star}(\tE)\succeq e^{-a}\kappa\,\Pi_{P^\star}$, and by definition of $t_k$, $\cH_{P^\star}(\t_0)\preceq t_k\,\cH^0_k$. Hence $v^\top\cH^0_kv\ge e^{-a}\kappa\xi/t_k$, and by~\eqref{eq:offline_proxy_comparison},
\[
    \lambda_k \;\ge\; e^{-a}\br{\frac{e^{-a}\kappa\,\xi}{t_k}-\varepsilon} \;\ge\; \frac{e^{-2a}\kappa\,\xi}{t_k}-\varepsilon.
\]
By Step~1, the warm-up condition~\eqref{eq:n_min}, which reads $\lambda_k\ge2R^2\beta_n^2(\delta)$, therefore holds as soon as
\[
    \frac{k}{\log(1+k)} \;\ge\; \frac{2d_U\,e^{2a}\br{2R^2\beta_n^2(\delta)+\varepsilon}}{\kappa\,\xi} \;=\; K.
\]
Since $\log(1+k)\le\log(2k)$, it suffices that $k\ge K\log(2k)$. As $\kappa\Pi_P\preceq\cH_P(\tE)\preceq\varepsilon\Pi_P$ gives $\kappa\le\varepsilon$, and $\xi\le1$, we have $K\ge2$, so Lemma~\ref{prop:linear_vs_log} shows that this holds for every $k\ge2K\log(4K)$. Since $n_\delta$ is the first round satisfying the warm-up condition, $n_\delta\le\roof{2K\log(4K)}$ whenever the right-hand side is at most $n$.
\end{proof}
While looser due to the less precise offline information $\cH^0_k$, this strategy retains a transfer rate that does not depend on $\abs{\cP}$, and provides a practical alternative when interaction is limited. The resulting offline algorithm is given in Appendix~\ref{app:algorithms}.

\section{Tractable warm-up length bound}
\label{app:computable_warmup}

The bound of Lemma~\ref{lemma:passive_learning} depends on $\tE$ through $\lambda_{\abs\cP}$ and $d_\star$. Both can be bounded using only the Hessians at $\t=0$, which only involve the uniform policy and are computed with the projectors (Appendix~\ref{app:algorithms}).

\begin{corollary}
\label{cor:computable_warmup}
Let $\bar\lambda\coloneq e^{-RB_\t}\lambda^U_{\min}\br{\sum_{P\in\cP}\cH_P(0)}$ and $\bar d_\star\coloneq d_U\,e^{RB_\t}H^2B_\phi^2\max_{P\in\cP}\lambda^{U_P}_{\min}\br{\cH_P(0)}^{-1}$, and let $\bar\beta_n^2(\delta)$ be defined as $\beta_n^2(\delta)$ with $\bar d_\star$ in place of $d_\star$. Under round-robin selection~\eqref{eq:roundrobin},
\[
    n_\delta \;\le\; \abs{\cP}\roof{\frac{2R^2\bar\beta_n^2(\delta)}{\bar\lambda}}.
\]
\end{corollary}
\begin{proof}
\emph{Step 1 (information after one sweep).} Since $\norm{\tE}_{\Pi_P}\le\norm{\tE}_2\le B_\t$ for every $P\in\cP$, item~1 of Lemma~\ref{lem:curvature_stability} with $\t_1=\tE$ and $\t_2=0$ gives
\[
    \cH_P(\tE)\succeq e^{-RB_\t}\,\cH_P(0).
\]
Summing over $\cP$ yields $\lambda_{\abs\cP}\ge\bar\lambda$.

\emph{Step 2 (effective dimension).} Since $\norm{\phi(\tau)}_2\le HB_\phi$, we have $\norm{\Sigma_P}_2\le\mathbb{E}\norm{\Pi_P\phi(\tau)}_2^2\le H^2B_\phi^2$. Moreover, Step~1 gives
\[
    \cH_P(\tE)^\dagger\preceq e^{RB_\t}\,\lambda^{U_P}_{\min}\br{\cH_P(0)}^{-1}\,\Pi_P,
\]
so that $d_\star\le\bar d_\star$. Since the threshold~\eqref{eq:n_min} is increasing in $d_\star$, it is at most $2R^2\bar\beta_n^2(\delta)$.

\emph{Step 3 (warm-up length).} Under round-robin selection, after $k$ rounds every $P\in\cP$ was selected at least $\floor{k/\abs\cP}$ times, so that
\[
    \lambda_k\;\ge\;\floor{k/\abs\cP}\,\lambda_{\abs\cP}\;\ge\;\floor{k/\abs\cP}\,\bar\lambda.
\]
By Step~2, the warm-up condition thus holds at round $k=\abs\cP\roof{2R^2\bar\beta_n^2(\delta)/\bar\lambda}$, which proves the claim.
\end{proof}

The factors $e^{RB_\t}$ make this bound loose, which is why we treat the warm-up length as a hyperparameter in practice (Section~\ref{sec:experiments}).

\section{Misspecified experts}
\label{apx:subopt_experts}

So far, we assumed that the expert is soft-optimal for $\tE$ under the learner's temperature $\eta$. We now only require its behavior to be explained by a nearby reward.

\begin{assumption}[Misspecified expert]
\label{as:misspecified_expert}
There exists $\tilde\tE\in\T$ with $R\norm{\tilde\tE-\tE}_{\Pi_U}\le 1$ such that, for every $P\in\cP$, the expert follows $\tilde\pi^\text{E}_P = \pi_P(\tilde\tE)$. We denote its suboptimality by
\[
    \varepsilon_\text{mis}\coloneq\max_{P\in\cP}\br{V^\star_P(\tE)-V^{\tilde\pi^\text{E}_P}_P(\tE)}.
\]
\end{assumption}

This covers a human demonstrator who is slightly mistaken about the task's objective, or whose level of randomness differs from the one the learner assumes. The latter simply amounts to a rescaled reward: an expert acting with an inverse temperature $\eta^\text{E}\neq\eta$ follows $\pi_P(\tilde\tE)$ with $\tilde\tE=(\eta^\text{E}/\eta)\,\tE$, since soft-optimal policies depend on the reward only through $\eta\t$, provided $(\eta^\text{E}/\eta)\norm{\tE}_2\le B_\t$. Its suboptimality then grows quadratically with the mismatch, 
\[
\varepsilon_\text{mis}\le(\eta^\text{E}/\eta-1)^2\max_{P\in\cP}\norm{\tE}^2_{\cH_P(\tE)}.
\]

Since the demonstrations are well-specified for $\tilde\tE$, all our results apply to $\tilde\tE$. Transferring to $\tE$ then only adds the expert's own suboptimality.

\begin{lemma}
\label{lemma:misspecified_transfer}
Let Assumption~\ref{as:misspecified_expert} hold, and let $d_\star$ and $n_\delta$ be defined with respect to the actual expert, i.e., at $\tilde\tE$. With probability at least $1-\delta$, it holds for all $P \in \cP$ and $n_\delta \le k \le n$ that
\[
    \subopt(\t_k,P) \;\lesssim\; g_k(P)\,\beta_k^2(\delta) \;+\; \varepsilon_\text{mis}.
\]
\end{lemma}
\begin{proof}
Let $w\coloneq\tilde\tE-\tE$ denote the misspecification and $u_k\coloneq\t_k-\tilde\tE$ the estimation error, so that $\t_k-\tE=u_k+w$, and write $\tilde\cH_P\coloneq\cH_P(\tilde\tE)$.

\emph{Step 1 (comparison of curvatures).} Since $U_P\subseteq U$, we have $R\norm{w}_{\Pi_P}\le R\norm{w}_{\Pi_U}\le 1$ for every $P\in\cP$, and item~1 of Lemma~\ref{lem:curvature_stability} gives
\[
    e^{-1}\cH_P(\tE)\preceq\tilde\cH_P\preceq e\,\cH_P(\tE).
\]
Hence, the information gain evaluated at $\tilde\tE$ is within a factor $e^2$ of $g_k(P)$, by the argument of Step~1 in the proof of Theorem~\ref{theo:active_learning}.

\emph{Step 2 (estimation error).} Since the demonstrations are generated by $\pi_P(\tilde\tE)$ with $\tilde\tE\in\T$, Lemma~\ref{thm:confidencenew} applies with $\tilde\tE$ in place of $\tE$ for all $n_\delta\le k\le n$. On its event, the proof of Lemma~\ref{lemma:transfer} and Step~1 yield
\[
    \norm{u_k}^2_{\tilde\cH_P}\le e^2 g_k(P)\,\beta_k^2(\delta)
    \qquad\text{and}\qquad
    R\norm{u_k}_{\Pi_P}\le 1,
\]
hence $R\norm{u_k+w}_{\Pi_P}\le 2$ for all $n_\delta\le k\le n$.

\emph{Step 3 (decomposition).} By Proposition~\ref{prop:subopt_remainder_form} and item~2 of Lemma~\ref{lem:curvature_stability},
\[
    \subopt(\t_k,P) \le \psi(2)\norm{u_k+w}^2_{\cH_P(\tE)} \le 2\psi(2)\br{\norm{u_k}^2_{\cH_P(\tE)}+\norm{w}^2_{\cH_P(\tE)}}.
\]
By Steps~1 and~2, the first term satisfies $\norm{u_k}^2_{\cH_P(\tE)}\le e\norm{u_k}^2_{\tilde\cH_P}\lesssim g_k(P)\,\beta_k^2(\delta)$. For the second, Proposition~\ref{prop:subopt_remainder_form} and item~2 of Lemma~\ref{lem:curvature_stability} with $\t_1=\tilde\tE$ and $\t_2=\tE$ give, since $R\norm{w}_{\Pi_P}\le 1$,
\[
    \varepsilon_\text{mis}\ge V^\star_P(\tE)-V^{\tilde\pi^\text{E}_P}_P(\tE)=\norm{w}^2_{G_P(\tilde\tE,\tE)}\ge\psi(-1)\norm{w}^2_{\cH_P(\tE)},
\]
which concludes the proof.
\end{proof}

Lemma~\ref{lemma:misspecified_transfer} is Lemma~\ref{lemma:transfer} plus a bias. Since $\t_k$ remains within distance $2/R$ of $\tE$ along every $U_P$ after the warm-up (Step~2), Theorem~\ref{theo:active_learning} carries over with larger constants and the same bias. In short, the learned reward transfers as well as the expert itself, up to the usual rate. No amount of demonstrations removes this bias.

The same holds for Theorem~\ref{theorem:transfer}, for any target dynamics $P$,
\[
    \subopt(\t_n,P) \;=\; \tilde\cO\br{\frac{d_\star\,d_U}{n\,m}} \;+\; \cO\br{\varepsilon_\text{mis} + g_\perp(P)\,B_\t^2},
\]
since the misspecification only enters the error along $U$, which the proof of Theorem~\ref{theorem:transfer} bounds through the dynamics in $\cP$, while the error along $U^\perp$ is bounded as before.

\begin{remark}[Unregularized experts]
\label{rem:unregularized}
Deterministic experts ($\eta\to\infty$) lie outside our framework, as entropy regularization is what makes the reward identifiable along each $U_P$~\citep{cao2021identifiability,schlaginhaufen2024towards}. They can be approximated with a large $\eta$, at the cost of constants growing with $\eta$, such as $R$.
\end{remark}

\section{Algorithms}
\label{app:algorithms}

\paragraph{Hessian estimation.}
Algorithm~\ref{alg:information} estimates the information $\cH_P(\t)$ of a dynamics $P$ from trajectories of its optimal policy $\pi_P(\t)$, using the feature advantages of Lemma~\ref{lemma:value_derivatives}, which require one policy evaluation of the features. Under deterministic transitions and initial state, the advantages reduce to centered trajectory features (Corollary~\ref{cor:deterministic_transitions}), so no policy evaluation is needed. Under stochastic dynamics, the advantages cannot be replaced by trajectory features, whose covariance also contains the dynamics noise $\Sigma^\text{dyn}_P$ (Lemma~\ref{lemma:feature_covariance}), which may lie outside $U_P$.

\paragraph{Projectors.}
Since $U_P$ does not depend on $\t$ (Proposition~\ref{lemma:image}), each projector is computed once, before any demonstration, from the information at $\t=0$, whose optimal policy $\pi_P(0)$ is uniform. We run Algorithm~\ref{alg:information} with $\pi_P(0)$ to obtain $\hat\cH_P(0)$, compute its eigendecomposition $\hat\cH_P(0)=\sum_{i=1}^d \hat\lambda_i\,\hat u_i\hat u_i^\top$, and set
\[
    \hat\Pi_P \coloneq \sum_{i:\,\hat\lambda_i>\varepsilon_{\lambda}} \hat u_i\hat u_i^\top.
\]
With exact Hessians, this recovers $\Pi_P=\cH_P(0)\cH_P(0)^\dagger$ for any $\varepsilon$ below the smallest positive eigenvalue of $\cH_P(0)$. The threshold $\varepsilon>0$ discards the small eigenvalues that sampling noise adds along unidentifiable directions. For our experiments, we fixed $\varepsilon_\lambda = 5 \cdot 10^{-4}$.

\paragraph{Two-phase active IRL.}
Algorithm~\ref{alg:online} details our two-phase approach. After computing the projectors, it selects dynamics round-robin~\eqref{eq:roundrobin} during the first $n_0$ rounds, observing $m$ expert trajectories in each, and then computes $\t_{n_0}$ with Algorithm~\ref{alg:multi_irl}. At every subsequent round $k$, it
\begin{enumerate}
    \item estimates $\cH_P(\t_{k-1})$ for each $P\in\cP$ with Algorithm~\ref{alg:information}, using the policies $\pi_P(\t_{k-1})$ returned by Algorithm~\ref{alg:multi_irl};
    \item selects $P_k$ by~\eqref{eq:active_selection_rule} and observes $m$ expert trajectories in $P_k$;
    \item updates $\t_k$ with Algorithm~\ref{alg:multi_irl} on all data collected so far.
\end{enumerate}

\paragraph{Offline active IRL.}
Algorithm~\ref{alg:offline} selects the whole sequence $\br{P_k}_{k=1}^n$ before observing the expert, using $\t_0=0$ in place of $\t_{k-1}$ and no warm-up. The required Hessians $\cH_P(0)$ are those already computed for the projectors. The expert is then observed on the selected dynamics, and the reward is estimated once with Algorithm~\ref{alg:multi_irl}.

\begin{algorithm}[h!]
\caption{Hessian estimation}
\label{alg:information}
\begin{algorithmic}[1]
\STATE \textbf{Input:} dynamics $P$, policy $\pi=\pi_P(\t)$, rollouts $N$
\STATE Set $\bar \Phi_{H+1}\coloneq 0$
\FOR{$h=H,\dots,1$}
\FOR{$a \in \cA, s \in \cS$}
\STATE $\Phi_h(s,a) \gets \phi(s,a) + \sum_{s'}T(s'\mid s,a)\,\bar\Phi_{h+1}(s')$
\STATE $\bar\Phi_h(s) \gets \sum_{a}\pi_h(a\mid s)\,\Phi_h(s,a)$
\ENDFOR
\ENDFOR
\STATE Sample $\tau_1,\dots,\tau_N$ with $\pi$ in $P$
\STATE $A(\tau_i) \gets \sum_{h=1}^H \br{\Phi_h(s_h,a_h) - \bar\Phi_h(s_h)}$ \COMMENT{Feature advantage}
\STATE \textbf{return:} $\hat\cH_P(\t) \gets \frac{\eta}{N}\sum_{i=1}^N A(\tau_i)A(\tau_i)^\top$ \COMMENT{Lemma~\ref{lemma:value_derivatives}}
\end{algorithmic}
\end{algorithm}

\begin{algorithm}[h!]
\caption{Projected min-max IRL with multiple dynamics}
\label{alg:multi_irl}
\begin{algorithmic}[1]
\STATE \textbf{Input:} expert data $\bc{(P_i,\hat\phi_i)}_{i=1}^k$, projectors $\bc{\hat\Pi_P}_{P\in\cP}$, step size $\alpha$, iterations $K$, rollouts $N$
\STATE \textbf{Initialize:} $\t^{(0)} \gets 0$
\FOR{$t=0,\dots,K-1$}
    \FOR{each distinct $P\in\bc{P_1,\dots,P_k}$}
        \STATE $\pi \gets \pi_P(\t^{(t)})$ \COMMENT{RL}
        \STATE $\bar\phi^{\,t}_P \gets \frac1N\sum_{j=1}^N \phi(\tau_j)$, with $\tau_j\sim\pi$ in $P$ \COMMENT{Rollouts}
    \ENDFOR
    \STATE $g_t \gets \sum_{i=1}^k \hat\Pi_{P_i}\br{\bar\phi^{\,t}_{P_i} - \hat\phi_i}$ \COMMENT{Gradient of~\eqref{eq:minmaxirl}}
    \STATE $\t^{(t+1)} \gets \Pi_\T\br{\t^{(t)} - \alpha g_t}$
\ENDFOR
\STATE $\t_k \gets \frac1K\sum_{t=0}^{K-1}\t^{(t)}$
\STATE \textbf{return:} $\t_k$, $\bc{\pi_P(\t_k)}_{P\in\cP}$
\end{algorithmic}
\end{algorithm}

\begin{algorithm}[h!]
\caption{Active IRL}
\label{alg:online}
\begin{algorithmic}[1]
\STATE \textbf{Input:} $\cP$, budget $n$, warm-up length $n_0\ge 0$, trajectories per round $m$, regularization $\varepsilon>0$, criterion $J$
\STATE \textbf{Initialize:} $\cD^\text{E}_0 \gets \emptyset$, $k(P) \gets 0$ for all $P\in\cP$, $\t_0\gets 0$
\STATE Compute $\hat\cH_P(0)$ and $\hat\Pi_P$ for all $P\in\cP$ from $\pi_P(0)$ \COMMENT{Algorithm~\ref{alg:information}}
\FOR{$k=1, \dots, n$}
    \IF[Warm-up]{$k \le n_0$}
        \STATE $P_{k} \in \argmin_{P\in\cP} k(P)$
    \ELSE[Active selection]
        \STATE Estimate $\hat\cH_P(\t_{k-1})$ for all $P \in \cP$ from $\pi_P(\t_{k-1})$ \COMMENT{Algorithm~\ref{alg:information}}
        \STATE $\hat\cH_{k-1} \gets \varepsilon\,\hat\Pi_U + \sum_{P\in\cP} k(P)\,\hat\cH_P(\t_{k-1})$
        \STATE $P_{k} \in \argmax_{P\in\cP}\; J \br{\hat\cH_{k-1}^{\dagger/2}\,\hat\cH_P(\t_{k-1})\,\hat\cH_{k-1}^{\dagger/2}}$
    \ENDIF
    \STATE $k(P_k) \gets k(P_k) + 1$
    \STATE $\cD^\text{E}_{k} \gets \cD^\text{E}_{k-1} \cup \bc{\tau_{k,j}}_{j=1}^m$, with $\tau_{k,j}$ generated by $\piE_{P_k}$ under $P_k$ \COMMENT{Observe expert}
    \IF{$k \ge n_0$}
        \STATE $\t_{k}, \bc{\pi_P(\t_k)}_{P\in\cP} \gets \operatorname{IRL}\br{\cD^\text{E}_{k}, \bc{\hat\Pi_P}_{P\in\cP}}$ \COMMENT{Algorithm~\ref{alg:multi_irl}}
    \ENDIF
\ENDFOR
\STATE \textbf{return:} $\t_n, \bc{\pi_P(\t_n)}_{P\in\cP}$
\end{algorithmic}
\end{algorithm}
\begin{algorithm}[t]
\caption{Offline Active IRL}
\label{alg:offline}
\begin{algorithmic}[1]
\STATE \textbf{Input:} $\cP$, budget $n$, trajectories per round $m$, regularization $\varepsilon>0$, criterion $J$
\STATE Compute $\hat\cH_P(0)$ and $\hat\Pi_P$ for all $P\in\cP$ from $\pi_P(0)$ \COMMENT{Algorithm~\ref{alg:information}}
\STATE $\cH^0_0 \gets \varepsilon_0\,\Pi_U$, with $\varepsilon \coloneq \eta H^2B_\phi^2$
\FOR{$k=1, \dots, n$}
    \STATE $P_k \in \argmax_{P\in\cP}\; J \br{(\cH^0_{k-1})^{\dagger/2}\,\hat\cH_P(0)\,(\cH^0_{k-1})^{\dagger/2}}$ \COMMENT{Offline active selection~\eqref{eq:active_selection_rule_offline}}
    \STATE $\cH^0_{k} \gets \cH^0_{k-1} + \hat\cH_{P_k}(0)$
\ENDFOR
\STATE $\cD^\text{E} \gets \bigcup_{k=1}^n\bc{\tau_{k,j}}_{j=1}^{m}$, with $\tau_{k,j}$ generated by $\piE_{P_k}$ under $P_k$ \COMMENT{Observe expert}
\STATE $\t_n, \bc{\pi_P(\t_n)}_{P\in\cP} \gets \operatorname{IRL}\br{\cD^\text{E}, \bc{\hat\Pi_P}_{P\in\cP}}$ \COMMENT{Algorithm~\ref{alg:multi_irl}}
\STATE \textbf{return:} $\t_n, \bc{\pi_P(\t_n)}_{P\in\cP}$
\end{algorithmic}
\end{algorithm}

\paragraph{Computational complexity.}
The main cost in our setting is the number of expert demonstrations, which our analysis minimizes. All other computations only require access to the dynamics in $\cP$, and we do not optimize them further. For completeness:
\begin{enumerate}
    \item \emph{Projectors (once):} $\abs{\cP}$ runs of Algorithm~\ref{alg:information}, each with one policy evaluation of $d$-dimensional features and one rollout batch, followed by an eigendecomposition of a $d\times d$ matrix.
    \item \emph{Information estimation (per active round):} $\abs{\cP}$ runs of Algorithm~\ref{alg:information} with the current policies, which Algorithm~\ref{alg:multi_irl} already returns, so no additional RL solve is needed.
    \item \emph{Dynamics selection (per active round):} one inverse square root of a $d\times d$ matrix and $\abs{\cP}$ evaluations of $J$, i.e., $\cO(\abs{\cP}\,d^3)$ operations.
    \item \emph{Reward estimation (per round):} $K$ gradient steps of Algorithm~\ref{alg:multi_irl}, each requiring one RL solve and one rollout batch per distinct dynamics selected so far, i.e., at most $\min(k,\abs{\cP})$, plus $\abs{\cP}$ final solves for the returned policies. In our finite-horizon tabular setting, each RL solve is a single backward pass of soft value iteration~\citep{geist2019theory}, i.e., $\cO(H\abs\cS^2\abs\cA)$ operations. Alternatively, each solve can be replaced by a single soft policy iteration step per reward update, which still converges~\citep{renard2024convergence} and avoids nested RL loops in larger problems.
\end{enumerate}
Memory is dominated by the $\abs{\cP}$ Hessians and projectors, i.e., $\cO(\abs{\cP}\,d^2)$. In comparison, the environment design approach of~\citet{buening2024environment} requires $\cO(k\abs{\cP})$ RL solves at round $k$ for selection alone.

\section{Additional experimental details and results}
\label{app:additional_experiment_details}

\subsection{Common Experimental Details}

All expert trajectories have length $H=100$. We set $\eta = 20$ and use $B_\phi = B_\theta = 1$ throughout all experiments.
Policies corresponding to every combination of dynamics and reward functions are computed using soft policy iteration.

In all experiments, the feature map $\phi$ is chosen as the indicator function over state--action pairs. Consequently, the reward function takes the form
\[
r_\theta(s,a)
=
\sum_{s',a'}
\mathds{1}\{s'=s,\; a'=a\}\,\theta_{s',a'}.
\]

Reported suboptimality is evaluated as the worst-case value over the considered set of dynamics $\mathcal{P}$, 
$\max_{P \in \mathcal{P}} \subopt(\theta_n, P)
$. 

\paragraph{Implementation details.}
The implementation is publicly available at
\href{https://anonymous.4open.science/r/transferable-active-irl}{https://anonymous.4open.science/r/transferable-active-irl}.

\paragraph{Compute resources.}
All experiments were conducted on a single \texttt{AMD EPYC 7302P} machine.

\subsection{Controlled sets of dynamics}
\label{section:principal}
\paragraph{Principal angles.}
Here we provide the details on principal angles, used for the first experimental setup. Intuitively, principal angles \citep{Galantai2004} can be seen as a natural generalization of angles between two lines or planes to higher dimensional subspaces. Formally we define them as follows.

\begin{definition}[Principal angles \citep{Galantai2004, schlaginhaufen2024towards}]
Let $\cV, \cW \subseteq \mathbb{R}^d$ be two subspaces. The principal angles $0 \leq \alpha_1(\cV, \cW) \leq \cdots \leq \alpha_m(\cV, \cW) := \alpha_{\max}(\cV, \cW) \leq \frac{\pi}{2}$ between $\cV$ and $\cW$ are defined recursively by
\begin{align*}
\cos\big(\alpha_i(\cV, \cW)\big)
= \max_{v \in \cV,\; w \in \cW} \ & \langle v, w \rangle \\
\text{s.t.} \quad 
& \|v\|_2 = \|w\|_2 = 1, \\
& \langle v, v_j \rangle = 0,\quad \langle w, w_j \rangle = 0, \quad j = 1, \dots, i-1,
\end{align*}
where $v_j \in \cV$ and $w_j \in \cW$ are maximizers corresponding to $\alpha_j(\cV, \cW)$.
\end{definition}

In the context of IRL, our work builds on the geometric perspective introduced by \citet{schlaginhaufen2024towards}. In particular, their Theorems~3.10 and~3.11 characterize reward transferability through the relative geometry of the unidentifiable subspaces associated with different transition dynamics, quantified via principal angles.
More precisely, Theorem~3.10 shows that when the \emph{second} principal angle between two unidentifiable subspaces $U_P^\perp$ and $U_{P'}^\perp$ is sufficiently large, observing the expert on both dynamics enables the recovery of rewards that transfer to arbitrary dynamics. Conversely, Theorem~3.11 establishes a complementary regime: when the \emph{largest} principal angle between $U_P$ and $U_{P'}$ is sufficiently small, a reward learned from demonstrations in one dynamics transfers reliably to the other.
Taken together, these results reveal a geometric trade-off. Large diversity between environments, captured by a large second principal angle, improves global identifiability and enables broad transfer. In contrast, strong similarity between dynamics, captured by a small maximal principal angle, guarantees local transferability of learned rewards across environments.

Let $\alpha_i(P, P')$ denote the $i$-th principal angle between the subspaces $U_P^\perp$ and $U_{P'}^\perp$, and let $\bar{\alpha}_i(P) = \frac{1}{\abs{\cP}}\sum_{P' \in \mathcal{P}} \alpha_i(P, P')$ be the average $i$-th principal angle of $P$ over the set $\mathcal{P}$.

\paragraph{Details.}

In our experiments, we constructed four classes of dynamics sets, each characterized by the geometry of their unidentifiable subspaces through principal-angle statistics. For a set of dynamics $\cP=\{P_1,\dots,P_{\abs{\cP}}\}$, we define
\[
\mu_i(\cP)
\coloneq
\frac{1}{\abs{\cP}}\sum_{P\in\cP}\bar\alpha_i(P),
\qquad
\sigma_i^2(\cP)
\coloneq
\frac{1}{\abs{\cP}}\sum_{P\in\cP}
\bigl(\bar\alpha_i(P)-\mu_i(\cP)\bigr)^2.
\]

All training dynamics are deterministic: every state-action pair $(s,a)$ maps to a single next state. For each seed, a set $\cP$ of $\abs{\cP}=10$ such dynamics is obtained by an evolutionary search over candidate sets, driven by a fitness defined on principal-angle statistics. A class is specified by a target angle rank $i\in\{2,\max\}$, a target angle $\alpha^\star$, and a heterogeneity weight $\lambda\ge 0$. For a candidate set $\cP$, let
\[
m_{\min}(\cP) \coloneq \frac{1}{\abs{\cP}}\sum_{P\in\cP}\min_{P'\neq P}\alpha_i(P,P'),
\qquad
m_{\max}(\cP) \coloneq \frac{1}{\abs{\cP}}\sum_{P\in\cP}\max_{P'\neq P}\alpha_i(P,P'),
\]
and let $h(\cP)\coloneq \sqrt{\pi/2}-\frac{1}{\abs{\cP}-1}\sum_{j=1}^{\abs{\cP}-1}\sqrt{\bar\alpha_{i,(j+1)}-\bar\alpha_{i,(j)}}$, where $\bar\alpha_{i,(1)}\le\dots\le\bar\alpha_{i,(\abs{\cP})}$ are the sorted average angles $\bar\alpha_i(P)$, $P\in\cP$. The fitness to minimize is
\[
f(\cP) \coloneq \max\Big\{\big(m_{\min}(\cP)-\alpha^\star\big)^2,\ \big(m_{\max}(\cP)-\alpha^\star\big)^2\Big\} + \lambda\, h(\cP).
\]
The first term drives both the closest and the farthest pair of every dynamics toward the target angle, so that transferability is uniform across pairs; the second term, when active, rewards large gaps between the sorted average angles, so that transferability is heterogeneous across the set. The four classes correspond to
\begin{itemize}
    \item Equivalent:  $i = \max, \alpha^\star = 0, \lambda = 0$;
    \item Similar: $ i = 2, \alpha^\star = 0, \lambda = 0$;
    \item Uniformly diverse: $i = 2, \alpha^\star = \pi/4, \lambda = 0$;
    \item Arbitrarily diverse:  $i = 2, \alpha^\star = \pi/3, \lambda = 10^4$.
\end{itemize}
The search maintains a population of $300$ candidate sets, initialized as copies of a set whose $\abs{\cP}$ dynamics are all equal to one random deterministic dynamics (next states drawn uniformly at random for every $(s,a)$). At every generation, the $30$ fittest candidates survive and $270$ offspring are generated, each a copy of a uniformly drawn survivor in which every transition is independently redirected to the initial state with probability $1/(\abs{\cS}\abs{\cA})$; when $i=2$, each dynamics of the offspring is additionally replaced by a fresh random deterministic dynamics with probability $0.1/\abs{\cP}$. The search runs for at most $500$ generations, stops early once $f(\cP)<10^{-5}$, and returns the fittest candidate, with dynamics indexed by increasing $\bar\alpha_i(P)$. Table~\ref{tab:set_stats} summarizes the statistics of the resulting dynamics sets, averaged over 20 runs.

\begin{table}[h]
\centering
\caption{Principal-angle statistics (in radian) of the controlled sets of dynamics, averaged over 20 runs.}
\label{tab:set_stats}
\begin{tabular}{lcc}
\toprule
Dynamics set
& $\mu_{2}(\cP)\pm \sigma_{2}(\cP)$
& $\mu_{\max}(\cP) \pm \sigma_{\max}(\cP)$ \\
\midrule
Equivalent dynamics
& $0.00\pm0.00$
& $0.00\pm0.00$ \\

Similar dynamics
& $0.00\pm0.00$
& $0.98\pm0.05$ \\

Uniformly diverse dynamics
& $0.03\pm0.03$
& $1.15\pm0.17$ \\

Arbitrarily diverse dynamics
& $0.57\pm0.11$
& $1.44\pm0.05$ \\
\bottomrule
\end{tabular}
\end{table}

For each seed, we evaluate the recovered reward on a set of $1000$ held-out stochastic dynamics $\{\tilde P_1,\dots,\tilde P_{1000}\}$, drawn independently of the training set $\cP$. For each $\tilde P_k$, a concentration vector $\beta\in\mathbb{R}^{\abs{\cS}}$ is drawn with $\log\beta_{s'}\sim\mathrm{Unif}[-2,\,0.1]$ independently for every next state $s'$, and every row $\tilde P_k(\cdot\mid s,a)$, $(s,a)\in\cS\times\cA$, is then sampled independently from $\mathrm{Dirichlet}(\beta)$. Since $\beta_{s'}\le e^{0.1}$, most rows concentrate their mass on a few next states, so the held-out dynamics are genuinely stochastic yet far from uniform, and none of them coincides with a deterministic training dynamics.

Regarding rewards, for each different seed, $\tE$ is uniformly sampled in $[0, 1]^d$, and then rescaled such that $\norm{r^\text{E}}_2 \le 1$. To estimate rewards, we run Algorithm~\ref{alg:multi_irl} for $100$ iterations, with $\alpha=0.02$.

\begin{figure}[t]
    \centering
    \includegraphics[width=\textwidth]{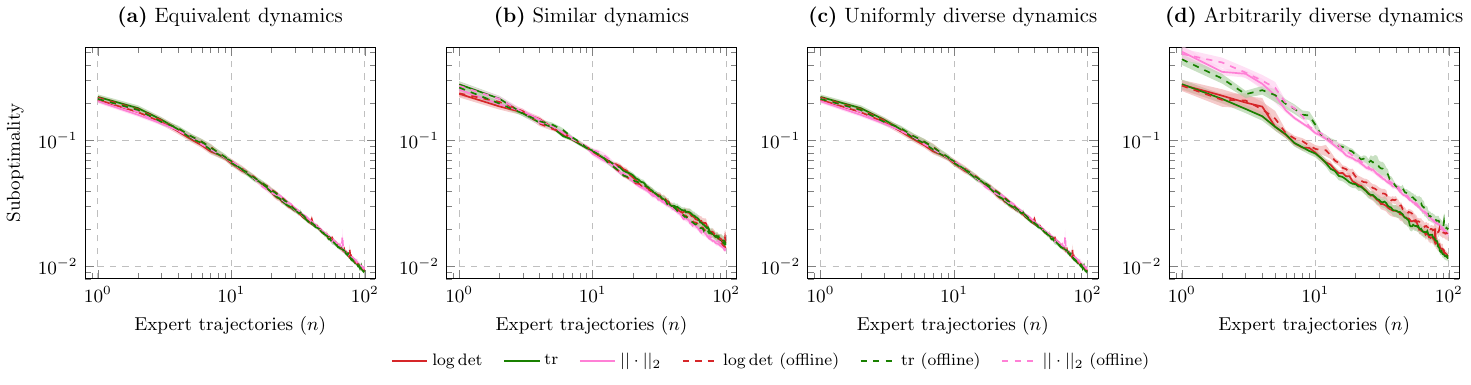}
\caption{
\looseness=-1
Worst-case suboptimality over the training set $\cP$ as a function of the number of observed expert trajectories on randomly generated MDPs, and of the criterion $J$ used to pick dynamics online~\eqref{eq:active_selection_rule} or offline~\eqref{eq:active_selection_rule_offline}. Each plot corresponds to a different structure of $\cP$.
}
\label{fig:controlled_set_apx_plots}
\end{figure}

\paragraph{Results.}
Figure~\ref{fig:controlled_set_apx_plots} reports the suboptimality attained by our methods when using either the $\log\det(I + \cdot)$, $\tr$, or $\norm{\cdot}_2$ criterion for the active selection objective $J$ defined in Equation~\eqref{eq:active_selection_rule}. In these experiments, all methods operate without a warm-up phase, meaning that dynamics are selected actively from the very first round. We additionally compare against the offline variant introduced in Section~\ref{app:offline_active_learning}.

We first observe that all methods achieve comparable performance when the dynamics set is uniformly distributed with respect to principal angles. Differences only emerge in the arbitrarily distributed setting, where the spectral norm criterion and the offline trace-based variant perform slightly worse, reaching approximately twice the suboptimality of the best-performing methods. Interestingly, the offline variant remains competitive in this setting.

Moreover, Table~\ref{tab:transferability_app} shows that the log-determinant and trace criteria provide the strongest transfer performance to previously unseen dynamics. In contrast, the offline variants generally exhibit weaker transferability overall, although the trace-based offline criterion performs best among the three offline alternatives.

\begin{table}[t]
\centering
\small
\caption{
Worst-case suboptimality on sets of \emph{held-out} stochastic dynamics after learning from $n=100$ expert trajectories collected on either uniformly or arbitrarily diverse sets of dynamics.
}
\setlength{\tabcolsep}{5pt}
\begin{tabular}{lccc}
\toprule
&& \multicolumn{2}{c}{Worst-case suboptimality ($\times 100$) on held-out dynamics} \\
\cmidrule(lr){3-4}
\textbf{criterion $J$\;\;}
& \textbf{Offline}
& \textbf{\;\;Uniformly diverse training set\;\;} 
& \textbf{\;\;Arbitrarily diverse training set\;\;} \\
\midrule
$\log\det(I + \cdot)$ & No & $\mathbf{02.86 \pm 0.02}$ & ${12.63 \pm 0.14}$ \\
$\tr$ & No & ${02.96 \pm 0.02}$ & $\mathbf{12.40 \pm 0.14}$ \\
$\norm{\cdot}_2$ & No & $03.02 \pm 0.19$ &  $15.23 \pm 0.17$ \\
\hdashline
$\log\det(I + \cdot)$ & Yes & $03.13 \pm 0.02$ &  $13.54 \pm 0.15$ \\
$\tr$ & Yes & $03.14 \pm 0.20$ &  $12.64 \pm 0.15$ \\
$\norm{\cdot}_2$ & Yes & $03.14 \pm 0.02$ &  $15.14 \pm 0.17$ \\
\bottomrule
\end{tabular}
\label{tab:transferability_app}
\end{table}

\begin{figure}[t]
    \centering
    \includegraphics[width=0.5\textwidth]{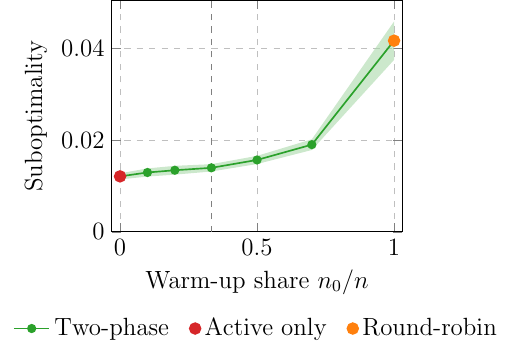}
\caption{
\looseness=-1
Worst-case suboptimality over the training set $\cP$ after $n=100$ rounds, as the warmup share $n_0/n$ increases. The vertical gray dashed line corresponds to the share of 1/3 used in the main paper.
}
\label{fig:warmup}
\end{figure}

Figure~\ref{fig:warmup} reports the worst-case suboptimality after $n=100$ rounds as a function of the warm-up share $n_0/n$, on the arbitrarily diverse dynamics sets, using the $\log\det(I+\cdot)$ criterion for the active phase. The two endpoints recover the Active only ($n_0=0$) and round-robin ($n_0=n$) selection rules. Suboptimality increases monotonically with the warm-up share. Actively selecting dynamics from the very first round is the best choice, and round-robin selection ends about three times worse. The degradation is however mild for short warm-ups: the share of $1/3$ used in the main paper costs roughly $15\%$ relative to no warm-up, and shares up to $1/2$ remain within $30\%$. Most of the gap opens once the warm-up exceeds half of the budget, where the active phase no longer has enough rounds to rebalance the counts toward the most informative dynamics. The warm-up therefore acts as a cheap safeguard against poor early Hessian estimates rather than as a source of performance.

\subsection{Random mazes}

\paragraph{Details.}
Mazes are $l \times l$ grids with $l=7$, i.e., $\abs{\cS}=49$ cells, with $\abs{\cA}=5$ actions corresponding to four moves and a waiting action. Transitions are deterministic, a move into a wall leaves the agent in place, and the initial state is the top-left cell. For each seed, we draw three goal cells $s_1,s_2,s_3$ uniformly among the non-initial cells, set $r^\text{E}(s_j, a_\text{wait}) \sim \cU[0,1]$, and $r^\text{E}(s,a)=0$ elsewhere. The reward vector is then rescaled to a uniformly random norm in $[0,1]$, so that $\norm{r^\text{E}}_2 \le 1$.

A set $\cP$ of $\abs{\cP}$ mazes is generated incrementally. The $i$-th maze ($i=0,\dots,\abs{\cP}-1$) starts from the walls of the $(i-1)$-th and is modified by a randomized wall-adding procedure that targets a shortest path from the initial state to the nearest goal of length $\ell_i := \big(\tfrac{i}{\abs{\cP}-1}\big)^2 l^2$, accepted within $\pm 5\%$. Walls are proposed uniformly at random and kept only if they do not shorten this path, so the path length is non decreasing in $i$. The procedure stops after $15{,}000$ proposals and returns the maze reached, so that $\ell_i$ acts as a target rather than a constraint. In practice, for $\abs{\cP}=100$, shortest paths range from $4$ to about $44$ steps, well below a Hamiltonian path. For odd $i$, the first wall that disconnects every goal from the initial state is kept, yielding an unsolvable maze. Roughly a third of each set is unsolvable, so that rewards must be inferred partly from dynamics in which the goals are unreachable.

For reward estimation, we run Algorithm~\ref{alg:multi_irl} with $100$ inner iterations per round using SGD with step size $0.02$.

\paragraph{Results.}

Beyond the results presented in the main paper, we additionally evaluate settings with dynamics sets of size $\abs{\cP}=5$ and $\abs{\cP}=20$ in order to study the impact of the number of candidate dynamics on the effectiveness of our active learning strategy.

\begin{figure}[t]
    \centering
    \includegraphics[width=0.9\textwidth]{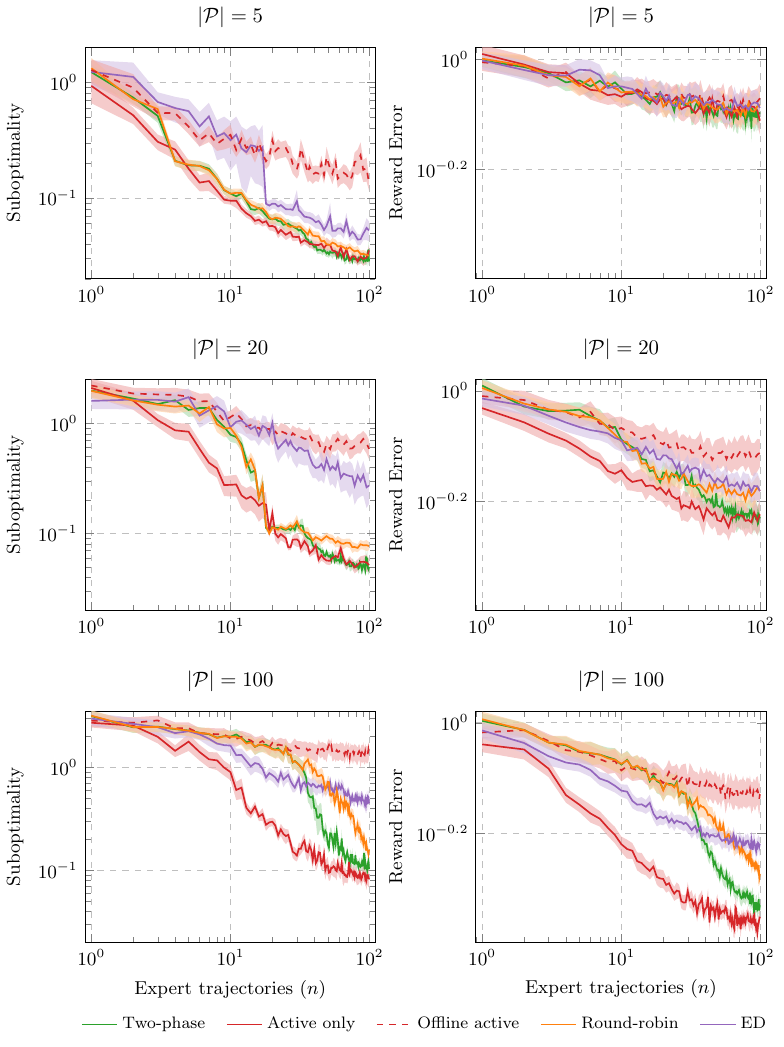}
    \caption{Worst-case suboptimality over the training set $\cP$ of randomly generated mazes, and the corresponding reward error, as a function of the number of observed expert trajectories, and of the size $\abs{\cP}$ of the dynamics set.}
    \label{fig:maze_plot_apx}
\end{figure}

Figure~\ref{fig:maze_plot_apx} reports both the suboptimality and reward estimation error achieved by our methods (using the log-determinant criterion), together with passive learning and  ED~\citep{buening2024environment}, for $\abs{\cP} \in \bc{5,20,100}$. Across all values of $\abs{\cP}$, our active selection rule (with and without warm-up) consistently outperforms the competing approaches. This suggests that actively selecting dynamics improves sample efficiency even when the set of candidate dynamics is relatively small.
However, the offline variant (see Section~\ref{app:offline_active_learning}) performs worse than passively selecting dynamics, hinting that in general, it is hard to entirely choose dynamics offline for expert demonstrations to learn the expert reward.

We also observe that the passive warm-up phase used in our two-phase method is not beneficial. In all tested settings, removing this warm-up phase leads to improved sample efficiency, with the gains becoming even more pronounced as the number of dynamics increases.

\end{document}